\documentclass{article}

\usepackage[main, final]{neurips_2026}

\usepackage[utf8]{inputenc} 
\usepackage[T1]{fontenc}    
\usepackage{hyperref}       
\usepackage{url}            
\usepackage{booktabs}       
\usepackage{amsfonts}       
\usepackage{nicefrac}       
\usepackage{microtype}      
\usepackage{xcolor}         
\usepackage{array}
\usepackage{amsmath}
\usepackage{amsthm}
\usepackage{mathrsfs}
\usepackage{bm}

\usepackage{booktabs}       
\usepackage{makecell}      

\usepackage{multirow}
\usepackage{caption}
\usepackage{subfigure}
\usepackage{graphicx}
\usepackage{color}
\usepackage{amssymb}
\usepackage{enumitem}
\usepackage{wrapfig}

\usepackage{pifont}    
\usepackage{colortbl}
\usepackage{xcolor} 
\usepackage{float}

\usepackage[most]{tcolorbox}   
\tcbuselibrary{skins,breakable}

\usepackage{algorithm}
\usepackage{algpseudocode}

\usepackage{tikz}

\usepackage{listings}

\usepackage{tabularx}

\usepackage{setspace}

\newtheorem{theorem}{Theorem}[section] 

\title{OpticalRAG: Pixel-Space Compression for Token-Efficient Retrieval-Augmented Generation}

\author{%
    Senhao Liu\textsuperscript{\rm 1}\thanks{Equal contribution.} \quad
    Yuheng Zhang\textsuperscript{\rm 1,2}\textsuperscript{$\ast$} \quad
    Chunyu Wei\textsuperscript{\rm 1}\thanks{Chunyu Wei is the corresponding author.} \quad
    Yueguo Chen\textsuperscript{\rm 1} \quad
    Xinran Zhang\textsuperscript{\rm 3} \\
    \textsuperscript{\rm 1}School of Information, Renmin University of China, Beijing, China \\
    \textsuperscript{\rm 2}ComindX \\
    \textsuperscript{\rm 3}Claremont McKenna College, Claremont, CA, USA \\
    \texttt{liusenhao@ruc.edu.cn}, \texttt{zhangyuheng@ruc.edu.cn}, \texttt{weicy15@icloud.com}, \\
    \texttt{chenyueguo@ruc.edu.cn}, \texttt{xzhang26@cmc.edu}
}

\begin{document}

\maketitle
\setcounter{footnote}{0}

\begin{abstract}
Retrieval-augmented generation (RAG) is bottlenecked by the token budget of large language models: feeding long documents into the context window is expensive, while aggressive textual compression discards fine-grained information that is irreversibly lost in a one-dimensional token sequence. We argue that this bottleneck is not fundamental to information density but to the choice of compression \emph{modality}. Modern visual encoders can pack the contents of a rendered text page into roughly a hundred visual tokens with little semantic loss, opening a new compression-fidelity tradeoff that purely textual methods cannot reach. We present \textbf{OpticalRAG}, a framework that renders documents as page images, encodes them with a frozen vision encoder, and distills the retrieved visual tokens into as few as $16$ injected tokens before passing them to the LLM. Two ingredients make this work without expensive cross-modal training: (i) \emph{Encoder-Bridged Retrieval}, which projects text-query states into the visual feature space of the frozen encoder, through a projection shared with the compressor, to retrieve visual tokens from a text query, with no contrastive training; and (ii) \emph{Query-Driven Distillation}, a lightweight cross-attention compressor that condenses the retrieved visual tokens into a small set of query-conditioned prefix tokens. On LongBench-v2, BAMBOO, and LooGLE-v2, OpticalRAG achieves the best compression-aware accuracy (CAP) on all three benchmarks while injecting only $16$ tokens, and improves the closed-book Qwen2.5-7B-Instruct backbone by $5.17$ points on LongBench-v2.~\footnote{Code is available at \url{https://github.com/brighterr/OpticalRAG}.}
\end{abstract}

\section{Introduction}

Large language models (LLMs) have demonstrated remarkable capabilities across knowledge-intensive tasks, yet they remain fundamentally constrained by finite context windows~\citep{DBLP:conf/nips/WeiHHWYWTC25}. When faced with long documents or large corpora, practitioners must either truncate the input or resort to retrieval-augmented generation (RAG), which introduces challenges around how retrieved content is represented and consumed by the LLM.

\begin{figure}[t]
    \centering
    \includegraphics[width=\linewidth]{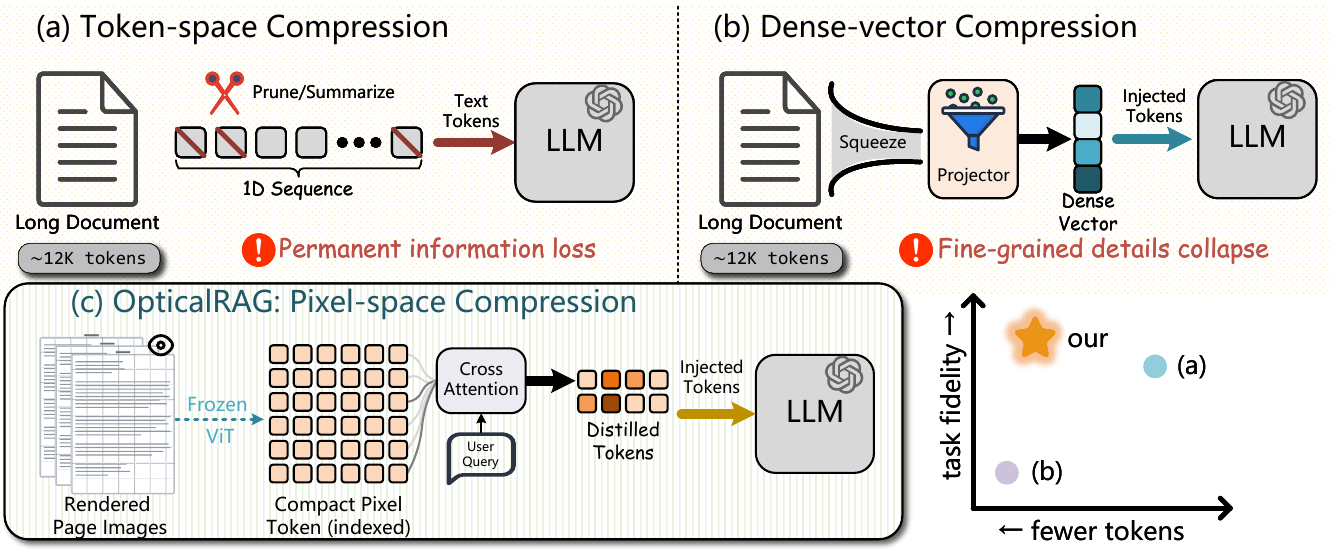}
    \caption{Illustration of OpticalRAG. \textbf{(a)} Token-space methods prune or summarize the 1D text sequence, causing permanent information loss at high compression ratios. \textbf{(b)} Dense-vector methods squeeze each document into a single embedding, collapsing fine-grained details. \textbf{(c)} Our \textbf{OpticalRAG} renders documents into page images and encodes them into compact visual tokens via a frozen ViT. At query time, a cross-attention module distills retrieved visual tokens into a small set of injected tokens conditioned on the user query.}
    \label{fig:paradigm}
\end{figure}

Existing approaches fall into two broad families (Figure~\ref{fig:paradigm}a-b). Textual compression methods, including prompt compression~\citep{DBLP:conf/acl/PanWJXLZLR0LZQ024}, extractive summarization~\citep{DBLP:conf/iclr/XuSC24}, and token pruning~\citep{DBLP:conf/emnlp/0001DGL23}, operate on text sequences, selectively discarding or merging tokens to fit within budgets. While effective at moderate compression ratios, they struggle to preserve fine-grained information at aggressive ratios because each removed token represents an irreversible loss in a one-dimensional sequence. A parallel line of work, exemplified by xRAG~\citep{DBLP:conf/nips/0002W00CWZ024}, attempts to bypass the token bottleneck by encoding entire documents into single dense embeddings injected into the LLM's representation space. Despite this progress, both families are \textit{modality-blind}: they ignore the structural redundancy inherent in how information is spatially laid out on a page. Single-embedding approaches collapse all document content into one vector, discarding fine-grained detail. Token-level methods treat text as a flat sequence, ignoring the rich 2D spatial relationships that humans naturally exploit when scanning a document.

We start from a simple observation: a single rendered page image, when processed by a modern visual encoder, can compress hundreds of text tokens into just a hundred visual tokens while preserving most semantic content. This motivates our core idea: instead of compressing text in the token space, compress it in the pixel space (Figure~\ref{fig:paradigm}c). Concretely, we render documents into page images, encode them into compact visual tokens for retrieval, and at query time distill the retrieved visual tokens through query-aware cross-attention into as few as 16 tokens injected directly into the LLM.

Realizing this idea effectively requires addressing two core challenges. \textbf{Challenge 1: Cross-Modal Alignment.} Documents are encoded as visual tokens, but user queries arrive as text. How can we build an effective retrieval bridge between text queries and visual document representations without costly end-to-end alignment training? \textbf{Challenge 2: Token-Budget Injection.} Even after retrieval, the retrieved visual tokens can still number in the hundreds or thousands. How can we distill only the query-relevant information into a minimal set of tokens?

To address \textit{Cross-Modal Alignment}, we introduce \textbf{Encoder-Bridged Retrieval}, which bridges the frozen visual encoder and the frozen LLM (Qwen2.5-7B-Instruct~\citep{DBLP:journals/corr/abs-2412-15115}): document chunks are rendered and encoded by the frozen visual encoder, while text queries are encoded by the LLM and projected into the same visual feature space via a lightweight projection shared with the compressor, requiring no contrastive training. 

To tackle \textit{Token-Budget Injection}, we propose \textbf{Query-Driven Distillation}, a lightweight cross-attention module that uses the query's semantic representation to condense retrieved visual tokens into a fixed small set of learned compression queries. Adapted from the Q-Former paradigm~\citep{DBLP:conf/icml/0008LSH23}, it dynamically routes information based on what the question demands, achieving compression from 256 retrieved visual tokens down to just 16 with minimal loss of task-relevant information.

\textbf{Contribution.} (1) We present \textbf{OpticalRAG}, a framework that reimagines document compression for RAG through visual encoding. By rendering text as images and operating entirely in the visual token space, we unlock a new compression-fidelity trade-off unavailable to purely textual methods. (2) Our two core techniques, \textit{Encoder-Bridged Retrieval} and \textit{Query-Driven Distillation}, address the cross-modal alignment and token-budget challenges respectively, enabling extreme compression while preserving task-critical information. (3) On LongBench-v2 (503 samples), OpticalRAG achieves an overall score of 35.79 with only 16 injected tokens, compared to 30.62 for the closed-book Qwen2.5-7B-Instruct backbone, a 5.17-point improvement while consuming orders of magnitude fewer tokens than textual retrieval and compression baselines.

\section{Related Work}

\paragraph{Context compression for RAG.}
Retrieval-augmented generation~\citep{DBLP:conf/icml/BorgeaudMHCRM0L22,DBLP:conf/iclr/QinWCW26,DBLP:conf/aaai/WeiQHWC26} enhances LLMs with external knowledge but introduces long contexts that are computationally expensive; we refer readers to \citet{DBLP:conf/naacl/LiLSC25} and \citet{DBLP:journals/corr/abs-2409-13385} for comprehensive surveys. Hard prompt compression methods reduce token counts by pruning or rewriting text, including self-information-based selection~\citep{DBLP:conf/emnlp/0001DGL23}, the LLMLingua family for query-aware compression~\citep{DBLP:conf/emnlp/JiangWLYQ23,DBLP:conf/acl/JiangWL0L0Q24}, and RAG-specific extractive compressors such as RECOMP~\citep{DBLP:conf/iclr/XuSC24}, EXIT~\citep{DBLP:conf/acl/HwangCJSHP25}, Provence~\citep{DBLP:conf/iclr/ChirkovaFNC25}, Refiner~\citep{DBLP:conf/emnlp/LiHLZH024}, and FiD-Light~\citep{DBLP:conf/sigir/HofstatterC0Z23}, with information-theoretic perspectives explored in~\citep{DBLP:conf/acl/0025FDGY0CCC024,DBLP:conf/emnlp/WangHTSYLFGC25}. Soft embedding methods instead compress contexts into continuous representations through Gist tokens~\citep{DBLP:conf/nips/Mu0G23}, ICAE~\citep{DBLP:conf/iclr/00010WWCW24}, AutoCompressors~\citep{DBLP:conf/emnlp/ChevalierWAC23}, Activation Beacon~\citep{DBLP:conf/iclr/Zhang0XSYD25}, 500xCompressor~\citep{DBLP:conf/acl/LiSC25}, and CEPE~\citep{DBLP:conf/acl/YenG024}, with RAG-tailored variants including PISCO~\citep{DBLP:conf/acl/LouisDC25}, CompAct~\citep{DBLP:conf/emnlp/YoonLHJK24}, Self-RAG~\citep{DBLP:conf/iclr/AsaiWWSH24}, LLoCO~\citep{DBLP:conf/emnlp/TanLPWZK0P24}, GritLM~\citep{DBLP:conf/iclr/MuennighoffS00W25}, and ECoRAG~\citep{DBLP:conf/acl/JeongKLH25}. Most relevant to our work, xRAG~\citep{DBLP:conf/nips/0002W00CWZ024} reinterprets dense retrieval embeddings as a cross-modal signal and injects each document as a single token, but the single-token bottleneck severely limits fine-grained information retention. KV-cache compression provides an orthogonal axis through token-eviction policies~\citep{DBLP:conf/nips/Zhang00CZC0TRBW23,DBLP:conf/iclr/XiaoTCHL24,DBLP:conf/nips/LiHYVLYCLC24,DBLP:conf/iclr/FuCAX0X25}, channel pruning~\citep{DBLP:conf/iclr/XuJD0LZSXS25}, quantization~\citep{DBLP:conf/icml/LiuYJZXBC024}, cross-layer merging~\citep{DBLP:conf/nips/LiuLPHHZ24}, and RAG-specific KV reuse~\citep{DBLP:conf/eurosys/YaoLLRCZD0J25,DBLP:conf/iclr/MaWL25,DBLP:conf/icml/HuHWWHZFCS025}. All these methods operate in the text token space; our work departs from this paradigm by compressing documents in the pixel space via visual encoding.

\paragraph{Visual document understanding.}
Visual document understanding has evolved from specialized Document AI models to general-purpose multimodal LLMs; see \citet{DBLP:conf/acl/DingLDJLSMLP26} for a recent survey. Early multimodal transformers including LayoutLMv3~\citep{DBLP:conf/mm/HuangL0LW22}, DocFormer~\citep{DBLP:conf/iccv/AppalarajuJKXM21,DBLP:conf/aaai/AppalarajuTDSZM24}, and UDoc~\citep{DBLP:conf/nips/GuKMZJBNS21} jointly modeled text, layout, and image patches, while the OCR-free paradigm pioneered by Donut~\citep{DBLP:conf/eccv/KimHYNPYHYHP22} and extended by Pix2Struct~\citep{DBLP:conf/icml/LeeJTH0EKSCT23} demonstrated that documents can be read directly from pixels. Modern multimodal LLMs further advance this direction through unified structure learning in DocOwl~1.5~\citep{DBLP:conf/emnlp/HuXYYZZZJHZ24}, the high-resolution compressor of DocOwl2~\citep{DBLP:conf/acl/Hu0ZY0ZJ0025}, OCR-augmented frozen VLMs in DocVLM~\citep{DBLP:conf/cvpr/NacsonAGBGKML25}, similarity-based token filtering in TextMonkey~\citep{DBLP:journals/pami/LiuYLLMZB26}, vision-language alignment in AlignVLM~\citep{DBLP:journals/corr/abs-2502-01341}, dual visual encoders in LLaVA-Read~\citep{DBLP:journals/corr/abs-2407-19185}, and hierarchical cross-attentive pooling in HVFA~\citep{DBLP:conf/nips/ParkCPH24}. For multi-page and multi-document scenarios, M3DocRAG~\citep{DBLP:journals/corr/abs-2411-04952} and SimpleDoc~\citep{DBLP:conf/emnlp/JainWZLDSWW25} embed document understanding in a RAG pipeline with visual retrievers, while Docopilot~\citep{DBLP:conf/cvpr/DuanCHWYSLHLLDW25} pursues native long-context understanding without retrieval. Most directly related is the emerging line of optical text compression: DeepSeek-OCR~\citep{DBLP:journals/corr/abs-2510-18234} maps rendered text into as few as 100 visual tokens per page while maintaining high OCR accuracy, its successor~\citep{DBLP:journals/corr/abs-2601-20552} introduces causal reordering of visual tokens, and GOT-OCR~\citep{DBLP:journals/corr/abs-2409-01704} proposes a unified end-to-end encoder-decoder for general OCR. On the visual token compression front, TokenCarve~\citep{DBLP:journals/corr/abs-2503-10501} and TCC~\citep{DBLP:journals/corr/abs-2407-14439} prune tokens via information-preservation and correlation criteria, PVC~\citep{DBLP:conf/cvpr/YangDZ0W00WLD25} introduces progressive compression across frames, and the Q-Former connector~\citep{DBLP:conf/emnlp/KimLPCOL24} provides the foundational query-token bottleneck architecture. Most closely, QueCC~\citep{DBLP:conf/iclr/LiGSK25} employs query-based convolutional cross-attention to compress visual tokens down to as few as one token, establishing scaling laws favoring extreme visual token reduction. Our work differs in that we do not aim to build a better document VLM; instead, we repurpose pretrained visual encoders as a compression and retrieval mechanism for general text-based RAG, adapting QueCC-style query-conditioned compression from a within-model setting to a cross-modal retrieval setting where the compressed content is retrieved documents rather than a currently observed image.

\section{Preliminary}

\textbf{Problem setting.}
We consider the standard retrieval-augmented generation scenario. Given a long document (or corpus) $\mathcal{D}$ consisting of $N$ text tokens and a user query $q$, the goal is to produce an answer $a$ by conditioning a frozen large language model $\mathcal{M}$ on a compressed representation of the relevant content in $\mathcal{D}$, subject to a strict token budget $B \ll N$. In the multiple-choice QA setting we primarily evaluate on, the answer $a^* \in \{A, B, C, D\}$ is selected by scoring the next-token logits of $\mathcal{M}$ over the candidate labels.

\textbf{Notation.}
We denote the visual encoder as $\mathcal{E}_{\text{vis}}$, the language model as $\mathcal{M}$ with hidden dimension $d_{\text{llm}}$, and the visual token dimension as $d_{\text{vis}}$. Bold lowercase letters (e.g., $\mathbf{v}$, $\mathbf{q}$) denote vectors, bold uppercase letters (e.g., $\mathbf{V}$, $\mathbf{Q}$) denote matrices, and calligraphic letters (e.g., $\mathcal{D}$, $\mathcal{M}$) denote sets or models. We use $[\,\cdot\,;\,\cdot\,]$ for sequence concatenation along the token dimension.

\section{Methodology}

\begin{figure}[t]
    \centering
    \includegraphics[width=\linewidth]{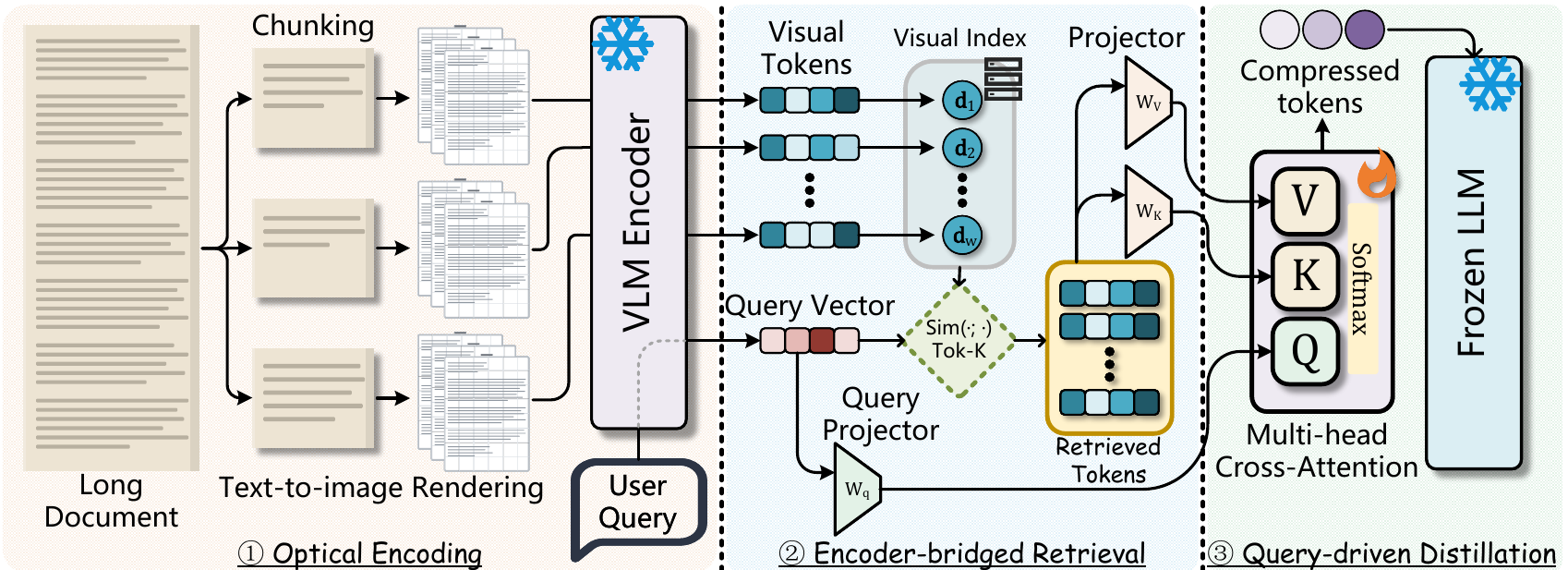}
    \caption{OpticalRAG framework overview. \ding{172}~\textbf{Optical Encoding}: document chunks are rendered as page images and encoded by a frozen VLM encoder into visual tokens; each visual token is indexed individually. \ding{173}~\textbf{Encoder-Bridged Retrieval}: the query is projected into visual space via $\mathbf{W}_q$ and matched to the index by inner product to retrieve top-$K$ visual tokens. \ding{174}~\textbf{Query-Driven Distillation}: a cross-attention module with learned compression queries distills retrieved visual tokens into a compact prefix fed into the frozen LLM.}
    \label{fig:framework}
\end{figure}

As illustrated in Figure~\ref{fig:framework}, OpticalRAG consists of three stages: (1) \emph{Optical Encoding} (\S\ref{sec:optical}), which renders text into page images and encodes them into compact visual tokens via a frozen visual encoder; (2) \emph{Encoder-Bridged Retrieval} (\S\ref{sec:retrieval}), which retrieves relevant visual tokens by projecting the query into the visual feature space; and (3) \emph{Query-Driven Distillation} (\S\ref{sec:distillation}), which compresses the retrieved visual tokens into a small set of query-conditioned tokens injected as a prefix into the frozen LLM. The visual encoder remains frozen; retrieval reuses the learned query projection from the distillation module.

\subsection{Optical Encoding}
\label{sec:optical}

The first stage converts a text document into a set of compact visual token sequences through text-to-image rendering followed by visual encoding.

\textbf{Text-to-image rendering.}
Given a document $\mathcal{D}$ with character length $|\mathcal{D}|$, we partition it into $W$ non-overlapping text windows by sequential splitting, where each window $\mathcal{D}_w$ contains at most $L_{\text{win}}$ characters: $\mathcal{D} \to \{\mathcal{D}_1, \ldots, \mathcal{D}_W\}$ with $|\mathcal{D}_w| \leq L_{\text{win}}$. Each window $\mathcal{D}_w$ is then rendered into a sequence of $P_w$ page images $\{\mathbf{I}_{w,1}, \ldots, \mathbf{I}_{w,P_w}\}$ at a fixed resolution of $H \times H$ pixels using a deterministic typesetting engine. The rendering is controlled by three parameters: font size $s_f$, margin width $s_m$, and line-height scale $s_l$, which together determine the information density per page. These parameters are hyperparameters of the compression pipeline. Rendering the windows yields $\sum_{w=1}^{W}P_w$ pages in total.

\textbf{Visual encoding.}
Each page image $\mathbf{I}_{w,p} \in \mathbb{R}^{H \times H \times 3}$ is passed through a frozen visual encoder $\mathcal{E}_{\text{vis}}$ to produce a fixed-length sequence of $T$ visual tokens, $\mathbf{V}_{w,p} = \mathcal{E}_{\text{vis}}(\mathbf{I}_{w,p}) \in \mathbb{R}^{T \times d_{\text{vis}}}$, where $T$ is the number of output patches and $d_{\text{vis}}$ is the visual embedding dimension. We adopt the DeepEncoder architecture~\citep{DBLP:journals/corr/abs-2510-18234}, which combines SAM and CLIP visual features to produce $T$ tokens per page. The concrete encoding and inference settings are provided in Appendix~\ref{app:impl:training}. All visual tokens from the same window are concatenated along the token dimension to form a window-level representation $\mathbf{V}_w = [\mathbf{V}_{w,1};\, \ldots;\, \mathbf{V}_{w,P_w}] \in \mathbb{R}^{(P_w \cdot T) \times d_{\text{vis}}}$. Across all windows, the complete visual representation of $\mathcal{D}$ is $\{\mathbf{V}_1, \ldots, \mathbf{V}_W\}$, containing $N_{\text{vis}} = T\sum_{w=1}^{W}P_w$ visual tokens. \emph{This is the core insight of OpticalRAG: by moving from 1D text sequences to 2D rendered images, we exploit the spatial compression ability of visual encoders to achieve a fundamentally different compression-fidelity trade-off.}

\subsection{Encoder-Bridged Retrieval}
\label{sec:retrieval}

Given the visual token representation of $\mathcal{D}$ and a text query $q$, we must identify the most relevant visual tokens for downstream answering. The key challenge is cross-modal alignment: queries are text while documents are visual tokens. We address this by learning a projection from language-model hidden states into the visual feature space.

\textbf{Document indexing.}
Each visual token $\mathbf{v}_i \in \mathbb{R}^{d_{\text{vis}}}$, $i=1,\ldots,N_{\text{vis}}$, is stored individually in an inner-product index for the current document, together with its original position. \emph{This indexing step is entirely offline and parameter-free: we reuse the frozen visual encoder's output without any additional training.}

\textbf{Query encoding.}
The query prompt $q$, including the question, candidate choices, and instructions, is tokenized and passed through the embedding layer and transformer blocks of the frozen language model $\mathcal{M}$ with a maximum prompt length of $L_q$ tokens to obtain its last-layer hidden states $\mathbf{H}_q \in \mathbb{R}^{n_q \times d_{\text{llm}}}$. Since $d_{\text{llm}} \neq d_{\text{vis}}$ in general, we first apply a learned linear projection $\mathbf{W}_q \in \mathbb{R}^{d_{\text{vis}} \times d_{\text{llm}}}$ to each hidden state, then max-pool over the token dimension to obtain $\tilde{\mathbf{h}}_q^\top = \text{MaxPool}(\mathbf{H}_q\mathbf{W}_q^\top)$. The projection $\mathbf{W}_q$ is part of the QueryDrivenCompressor module and is jointly trained with the distillation stage.

\textbf{Retrieval.}
We retrieve the top-$K$ visual tokens by inner product: $\mathcal{R}(q) = \operatorname*{top\text{-}K}_{i \in \{1,\ldots,N_{\text{vis}}\}} \tilde{\mathbf{h}}_q^\top \mathbf{v}_i$, retaining all candidates if fewer than $K$ are available.

The selected tokens are sorted by their original positions before being passed to the query-driven distillation module.

\subsection{Query-Driven Distillation}
\label{sec:distillation}

Retrieval selects $S=\min(K,N_{\text{vis}})$ visual tokens. The Query-Driven Distillation module compresses this selected sequence into $M$ query-conditioned visual-prefix tokens, reducing the amount of document-derived context passed to the frozen LLM.

\textbf{Visual token preparation.}
We first stack all retrieved visual tokens into a single sequence $\mathbf{V}_{\text{all}} = [\mathbf{v}_{r_1}^\top;\, \ldots;\, \mathbf{v}_{r_S}^\top] \in \mathbb{R}^{S \times d_{\text{vis}}}$, where $\{r_1, \ldots, r_S\} = \mathcal{R}(q)$, $r_1<\cdots<r_S$, and $S=\min(K,N_{\text{vis}})$ is the number of selected visual tokens.

\textbf{Query-global fusion.}
To inform the compression with query semantics, we use the same projected and pooled query representation as in retrieval: $\mathbf{q}_{\text{global}} = \tilde{\mathbf{h}}_q \in \mathbb{R}^{d_{\text{vis}}}$. The query-global vector is broadcast and added to each visual token before downsampling to produce query-fused visual representations $\mathbf{V}_{\text{fused}} = \mathbf{V}_{\text{all}} + \mathbf{1}\mathbf{q}_{\text{global}}^\top \in \mathbb{R}^{S \times d_{\text{vis}}}$, where $\mathbf{1} \in \mathbb{R}^{S}$ is the all-ones vector. This step introduces query information into the token representations. Query-dependent attention weights are also enabled by conditioning the compression queries below; token-side addition alone does not guarantee query-dependent selection (Appendix~\ref{app:qformer}).

When downsampling is enabled and $S$ is a perfect square, the implementation reshapes the fused sequence into a square grid and applies a depthwise convolution with kernel size and stride $s$. This operation requires $\sqrt{S}\ge s$; for valid square inputs,
\begin{equation}
    \hat{\mathbf{V}}_{\text{fused}} = \text{DWConv}_{\downarrow s}(\mathbf{V}_{\text{fused}}) \in \mathbb{R}^{S' \times d_{\text{vis}}}, \quad S' = \left\lfloor\frac{\sqrt{S}}{s}\right\rfloor^2.
    \label{eq:dsconv}
\end{equation}
For non-square sequence lengths or when downsampling is disabled, $\hat{\mathbf{V}}_{\text{fused}}=\mathbf{V}_{\text{fused}}$, with $S'=S$. The implementation does not add a separate guard for square inputs whose side length is smaller than the kernel. The grid is formed from the ordered retrieved sequence and does not necessarily preserve adjacency on the original pages.

\textbf{Learned compression queries.}
We introduce $M$ learnable query tokens $\mathbf{Z} \in \mathbb{R}^{M \times d_{\text{vis}}}$, randomly initialized and trained end-to-end. These serve as the ``information sinks'' that absorb compressed knowledge from the visual tokens. We initialize $\mathbf{Z}^{(0)} = \mathbf{Z} + \mathbf{1}_M\mathbf{q}_{\text{global}}^\top$, where $\mathbf{1}_M$ is the all-ones vector of length $M$, and form keys and values using learned projections: $\mathbf{K}_{\text{vis}}=\hat{\mathbf{V}}_{\text{fused}}\mathbf{W}_K^\top$ and $\mathbf{V}_{\text{val}}=\hat{\mathbf{V}}_{\text{fused}}\mathbf{W}_V^\top$, with $\mathbf{W}_K,\mathbf{W}_V\in\mathbb{R}^{d_{\text{vis}}\times d_{\text{vis}}}$. The compression is performed by a stack of $L$ multi-head cross-attention layers:
\begin{equation}
    \mathbf{Z}^{(\ell)} = \mathbf{Z}^{(\ell-1)} + \text{MHA}^{(\ell)}\!\left(\mathbf{Z}^{(\ell-1)},\; \mathbf{K}_{\text{vis}},\; \mathbf{V}_{\text{val}}\right), \quad \ell = 1, \ldots, L,
\end{equation}
where $\text{MHA}^{(\ell)}(\mathbf{Q}, \mathbf{K}, \mathbf{V})$ denotes multi-head cross-attention at layer $\ell$ with $\mathbf{Q}$ as queries and $\mathbf{K}, \mathbf{V}$ as keys and values, using $n_h$ attention heads. The output $\mathbf{Z}_{\text{out}}=\text{LN}(\mathbf{Z}^{(L)}) \in \mathbb{R}^{M \times d_{\text{vis}}}$ contains $M$ compressed tokens trained to retain query-relevant information from the retrieved visual tokens. \emph{These $S'$ query-fused visual tokens are distilled into $M$ compressed tokens through learned cross-attention, with the query guiding which information to preserve.}

\textbf{Bridge to LLM space.}
The compressed tokens $\mathbf{Z}_{\text{out}}$ reside in the visual embedding space ($d_{\text{vis}}$) and must be projected into the LLM's input space ($d_{\text{llm}}$). We employ a two-layer MLP with LayerNorm:
\begin{equation}
    \begin{aligned}
    \mathbf{U} &= \operatorname{GELU}\!\left(
      \mathbf{Z}_{\text{out}}\mathbf{W}_1^\top+\mathbf{1}_M\mathbf{b}_1^\top\right),\\
    \mathbf{P} &= \operatorname{LN}\!\left(
      \operatorname{Dropout}_{p_{\mathrm{drop}}}(\mathbf{U})\mathbf{W}_2^\top
      +\mathbf{1}_M\mathbf{b}_2^\top\right).
    \end{aligned}
    \label{eq:bridge}
\end{equation}
Here $\mathbf{W}_1\in\mathbb{R}^{d_h\times d_{\text{vis}}}$,
$\mathbf{W}_2\in\mathbb{R}^{d_{\text{llm}}\times d_h}$,
$\mathbf{b}_1\in\mathbb{R}^{d_h}$, and
$\mathbf{b}_2\in\mathbb{R}^{d_{\text{llm}}}$ are learned parameters;
$\mathbf{1}_M$ broadcasts the biases across prefix positions.
For the chosen backbones, $d_{\text{vis}}=1024$ and $d_{\text{llm}}=3584$; $d_h$ and $p_{\mathrm{drop}}$ denote the bridge hidden width and dropout probability.
Dropout is active during training and disabled during evaluation, and
$\mathbf{P}\in\mathbb{R}^{M\times d_{\text{llm}}}$.
The resulting $\mathbf{P}$ serves as a soft prefix prepended to the query's text token embeddings: $\mathbf{X}_{\text{input}} = [\mathbf{P};\; \text{Embed}_{\mathcal{M}}(q)] \in \mathbb{R}^{(M + n_q) \times d_{\text{llm}}}$, where $\text{Embed}_{\mathcal{M}}(q)$ denotes the token embeddings of the query prompt obtained from the frozen $\mathcal{M}$. Here $n_q$ is the prompt length after tokenization and truncation. The budget $M$ counts only the visual prefix; the answer-scoring forward pass receives $M+n_q$ input embedding positions.

\textbf{Answer prediction.}
The composite input $\mathbf{X}_{\text{input}}$ is fed into the frozen language model $\mathcal{M}$. For multiple-choice QA, the model selects an answer from the candidate options associated with each question.

\subsection{Training}

Only the QueryDrivenCompressor (including the projection $\mathbf{W}_q$, the depthwise convolution in Eq.~\ref{eq:dsconv}, the $L$-layer cross-attention, and the learnable queries $\mathbf{Z}$) and the VisualToLLMBridge (Eq.~\ref{eq:bridge}) are trained. The visual encoder $\mathcal{E}_{\text{vis}}$ and the language model $\mathcal{M}$ remain entirely frozen throughout.

\textbf{Stage 1: Reconstruction pretraining.}
We first pretrain the compressor and bridge on a text reconstruction objective. Given a document chunk rendered into visual tokens, the model must reconstruct the original text from the compressed prefix tokens. The training data consists of passages sampled from Wikipedia and C4. The loss is the standard next-token prediction loss:
\begin{equation}
    \mathcal{L}_{\text{recon}} = -\sum_{t=1}^{n} \log p_{\mathcal{M}}(x_t \mid \mathbf{P},\, x_{<t}),
\end{equation}
where $x_1, \ldots, x_n$ are the tokens of the original text and $\mathbf{P}$ is the compressed prefix. \emph{This stage teaches the compressor to faithfully encode document content into the prefix tokens, establishing a strong initialization before task-specific adaptation.}

\textbf{Stage 2: Task-aware fine-tuning.}
We then fine-tune on reading comprehension datasets (SQuAD~v2, DROP, CoQA). The loss is computed over the answer:
\begin{equation}
    \mathcal{L}_{\text{task}} = -\sum_{t=1}^{n_a} \log p_{\mathcal{M}}(a_t \mid \mathbf{P},\, q,\, a_{<t}),
\end{equation}
where $a_1, \ldots, a_{n_a}$ are the answer tokens. We use early stopping based on a held-out development set, selecting the checkpoint with the best development performance. Training hyperparameters and data split sizes are reported in Table~\ref{tab:app:training-config}. \emph{The two-stage strategy decouples content-preservation learning from task-specific reasoning, allowing the compressor to first learn ``what to remember'' and then ``what matters for answering.''}
\section{Theoretical Analysis}
\label{sec:theory}

We show that Query-Driven Distillation can compress $S'$ retrieved visual tokens into $M$ prefix tokens with arbitrarily small loss in task-relevant information (exactly lossless in the hard-attention limit), provided $M$ is at least the intrinsic query-conditioned rank $r$ of the answer distribution.

\begin{theorem}[Lossless Query-Conditioned Compression]
\label{thm:lossless}
Let $\hat{\mathbf{V}}_{\text{fused}} \in \mathbb{R}^{S' \times d_{\text{vis}}}$ with rows $\hat{\mathbf{v}}_s$ satisfying $\|\hat{\mathbf{v}}_s\|_2 \leq R$, and $\mathbf{q}_{\text{global}} \in \mathbb{R}^{d_{\text{vis}}}$. Suppose there exist $\mathbf{U} \in \mathbb{R}^{d_{\text{vis}} \times r}$ with orthonormal columns, probe vectors $\mathbf{z}^\star_1, \ldots, \mathbf{z}^\star_r \in \mathbb{R}^{d_{\text{vis}}}$, a margin $\gamma > 0$, and a function $g$ that is $L_g$-Lipschitz in its first argument (from the Frobenius to the Euclidean norm), such that for every input: (i) for each $j$, the evidence position $\sigma(j) := \arg\max_{s} (\mathbf{z}^\star_j + \mathbf{q}_{\text{global}})^\top \hat{\mathbf{v}}_s$ satisfies $(\mathbf{z}^\star_j + \mathbf{q}_{\text{global}})^\top (\hat{\mathbf{v}}_{\sigma(j)} - \hat{\mathbf{v}}_s) \geq \gamma$ for all $s \neq \sigma(j)$; and (ii)
\begin{equation}
    p\!\left(a^\star \mid \hat{\mathbf{V}}_{\text{fused}}, \mathbf{q}_{\text{global}}\right) = g\!\left(\mathbf{U}^\top \bigl[\hat{\mathbf{v}}_{\sigma(1)}, \ldots, \hat{\mathbf{v}}_{\sigma(r)}\bigr],\; \mathbf{q}_{\text{global}}\right).
    \label{eq:sufficiency}
\end{equation}
Then for any $M \geq r$ and $\varepsilon > 0$, there exist a single-layer $\mathrm{MHA}$, learnable queries $\mathbf{Z} \in \mathbb{R}^{M \times d_{\text{vis}}}$ conditioned as $\mathbf{Z}^{(0)} = \mathbf{Z} + \mathbf{1}_M \mathbf{q}_{\text{global}}^\top$ (Section~\ref{sec:distillation}), and a readout $\Phi$ such that, uniformly over all such inputs,
\begin{equation}
    \Bigl\|\Phi\!\left(\mathrm{MHA}\!\left(\mathbf{Z}^{(0)},\, \hat{\mathbf{V}}_{\text{fused}},\, \hat{\mathbf{V}}_{\text{fused}}\right),\, \mathbf{q}_{\text{global}}\right) - p\!\left(a^\star \mid \hat{\mathbf{V}}_{\text{fused}},\, \mathbf{q}_{\text{global}}\right)\Bigr\|_2 \leq \varepsilon,
    \label{eq:lossless}
\end{equation}
with equality of the two distributions in the limit of the attention temperature $\tau \to \infty$.
\end{theorem}

\begin{proof}
Write $\mathbf{u}_1, \ldots, \mathbf{u}_r$ for the columns of $\mathbf{U}$. By assumption~\eqref{eq:sufficiency}, it suffices to construct $\mathrm{MHA}$ and $\Phi$ such that the rows of $\mathbf{Z}^{(1)}$ encode the matrix $\mathbf{A} := \mathbf{U}^\top [\hat{\mathbf{v}}_{\sigma(1)}, \ldots, \hat{\mathbf{v}}_{\sigma(r)}] \in \mathbb{R}^{r \times r}$, since $\Phi$ can then apply $g$ to obtain the answer distribution.

Choose $n_h \geq r$ (allowing each head full width $d_{\text{vis}}$) and, without loss of generality, take $M = r$; additional queries and heads are ignored by $\Phi$. For each head $i \in \{1,\ldots,r\}$, define
\begin{equation}
    \mathbf{W}^{(i)}_K = \mathbf{I}_{d_{\text{vis}}}, \qquad \mathbf{W}^{(i)}_Q = \tau \, \mathbf{I}_{d_{\text{vis}}}, \qquad \mathbf{W}^{(i)}_V = \mathbf{u}_i \mathbf{u}_i^\top,
\end{equation}
with the $j$-th learnable query set to the fixed probe $\mathbf{z}_j = \mathbf{z}^\star_j$, so that the $j$-th row of $\mathbf{Z}^{(0)}$ is $\mathbf{z}^\star_j + \mathbf{q}_{\text{global}}$. The attention logit between query $j$ and token $s$ is
\begin{equation}
    \alpha_{j,s} = \tau \, (\mathbf{z}^\star_j + \mathbf{q}_{\text{global}})^\top \hat{\mathbf{v}}_s,
\end{equation}
which by (i) attains its unique maximum at $s = \sigma(j)$ with margin $\tau\gamma$. Hence the softmax weights $\beta_{j,s}$ place total mass at most $(S'-1)e^{-\tau\gamma}$ on $s \neq \sigma(j)$, and the head-$i$ output satisfies
\begin{equation}
    \mathbf{o}^{(i)}_j = \sum_{s} \beta_{j,s} \bigl(\mathbf{u}_i^\top \hat{\mathbf{v}}_s\bigr) \mathbf{u}_i, \qquad \bigl|\mathbf{u}_i^\top \mathbf{o}^{(i)}_j - \mathbf{u}_i^\top \hat{\mathbf{v}}_{\sigma(j)}\bigr| \leq 2R(S'-1)e^{-\tau\gamma}.
\end{equation}
Concatenating all $r$ heads and reading off the coefficients along the orthonormal basis $\mathbf{u}_1, \ldots, \mathbf{u}_r$ yields an estimate $\hat{\mathbf{A}}[:, j]$ of $\mathbf{A}[:, j] = \mathbf{U}^\top \hat{\mathbf{v}}_{\sigma(j)} \in \mathbb{R}^r$ for each query $j$, so that $\|\hat{\mathbf{A}} - \mathbf{A}\|_F \leq 2rR(S'-1)e^{-\tau\gamma}$.
Let $\Phi$ return $g(\hat{\mathbf{A}}, \mathbf{q}_{\text{global}})$. By~\eqref{eq:sufficiency} and the Lipschitz continuity of $g$, the error in~\eqref{eq:lossless} is at most $2L_g rR(S'-1)e^{-\tau\gamma}$, which is at most $\varepsilon$ for $\tau \geq \gamma^{-1}\log\bigl(2L_g rR(S'-1)/\varepsilon\bigr)$ and vanishes as $\tau \to \infty$, establishing~\eqref{eq:lossless}. \hfill$\square$
\end{proof}

The theorem gives an answer to ``how small can $M$ be?'': $M$ need only match the query-conditioned intrinsic rank $r$, not the size of the visual token pool. This is consistent with $M{=}16$ sufficing in practice (multiple-choice questions rarely require more than a handful of semantic factors), and explains why query-conditioning of the compression queries is essential (it lets fixed probes select question-dependent evidence) and why a single cross-attention layer is theoretically sufficient under these assumptions (deeper stacks aid optimization rather than being required for this construction).

\definecolor{headerblue}{RGB}{31,73,125}
\definecolor{oursblue}{RGB}{219,229,241}
\definecolor{accentline}{RGB}{31,73,125}
\newcommand{\best}[1]{\textbf{#1}}
\newcommand{\sbest}[1]{\underline{#1}}

\section{Experiments}
\label{sec:experiments}

\textbf{Evaluation datasets.}
To ensure a unified objective-question evaluation setting, we evaluate all methods on multiple-choice data only. We use the full LongBench-v2 benchmark~\citep{DBLP:conf/acl/BaiTZ0WLCX0D0L25} with 503 challenging samples covering diverse long-context reasoning scenarios, the choice-question subsets of BAMBOO~\citep{DBLP:conf/coling/DongTLZW24} , and the choice subsets of LooGLE-v2~\citep{DBLP:conf/nips/HeWLLZ25} .

\textbf{Evaluation protocol.}
We evaluate each method by task score $S$ and inference-time token budget $B$. To account for compression efficiency, we report $\mathrm{CAP}$, which combines performance with a compression multiplier. Let $S_0$ be the closed-book backbone score and $B_0=12\mathrm{k}$ be its token budget. For a method with score $S$ and budget $B$, we define $m(B)=((B_0+1)/(B+1))^{\delta}$ with $\delta=0.05$, following the inference-time VLM scaling trend for OCR-style document tasks~\citep{DBLP:conf/iclr/LiGSK25}. The final score is $\mathrm{CAP}(S,B)=S\cdot m(B)$ if $S\geq S_0$, and $\mathrm{CAP}(S,B)=S$ otherwise. This score-floor follows the Resilience-Rate principle~\citep{DBLP:conf/nips/0002W00CWZ024}, granting compression credit only when the method does not underperform the closed-book backbone.

\textbf{Baselines.}
We compare against several representative baselines under the same backbone, \texttt{Qwen2.5-7B-Instruct}: a closed-book \textit{Backbone} that answers without long context (measuring parametric knowledge and choice bias); \textit{Standard RAG} that retrieves textual chunks and feeds them to the backbone; and the compression baselines xRAG~\citep{DBLP:conf/nips/0002W00CWZ024}, LLMLingua-2~\citep{DBLP:conf/acl/PanWJXLZLR0LZQ024}, Selective Context~\citep{DBLP:conf/emnlp/0001DGL23}, and RECOMP~\citep{DBLP:conf/iclr/XuSC24}, which respectively use retrieval embedding injection, prompt compression, information-based filtering, and evidence condensation.

\subsection{Main Results}
\label{sec:main_results}

\begin{table}[t]
  \centering
  \scriptsize
  \setlength{\tabcolsep}{3pt}
  \renewcommand{\arraystretch}{1.15}
  \caption{Main results on LongBench-v2, Bamboo, and LooGLE-v2.
           }
  \label{tab:main_results}
  \arrayrulecolor{accentline}
  \resizebox{\linewidth}{!}{%
  \begin{tabular}{ll cccccc c}
    \toprule
    \rowcolor{headerblue}
    \multicolumn{2}{l}{\textcolor{white}{}}
    & \textcolor{white}{\textsc{Backbone}}
    & \textcolor{white}{\textsc{Std.\ RAG}}
    & \textcolor{white}{\textsc{Sel.\ Context}}
    & \textcolor{white}{\textsc{LLMLingua-v2}}
    & \textcolor{white}{\textsc{Xrag}}
    & \textcolor{white}{\textsc{RECOMP}}
    & \cellcolor{oursblue}\textcolor{headerblue}{\textsc{Ours}} \\
    \midrule
    \multirow{3}{*}{\textbf{LongBench-v2}}
      & \textit{Score}
        & 30.62 & 31.41 & \sbest{32.40} & 28.23 & 26.24 & 32.01
        & \cellcolor{oursblue}\best{35.79} \\
      & \textit{Len}
        & 0 & 1622.68 & 5511.73 & 2669.19 & 1 & 6953.93
        & \cellcolor{oursblue}16 \\
      & \textit{CAP} $\uparrow$
        & 30.62 & 34.71 & 33.69 & 28.23 & 26.24 & \sbest{32.90}
        & \cellcolor{oursblue}\best{49.68} \\
    \midrule
    \multirow{3}{*}{\textbf{Bamboo}}
      & \textit{Score}
        & 56.75 & 69.50 & 70.00 & \sbest{73.50} & 51.00 & \best{75.50}
        & \cellcolor{oursblue}61.00 \\
      & \textit{Len}
        & 0 & 814.22 & 3141.29 & 1667.27 & 1 & 2415.46
        & \cellcolor{oursblue}16 \\
      & \textit{CAP} $\uparrow$
        & 56.75 & 79.50 & 74.85 & 81.12 & 51.00 & \sbest{81.80}
        & \cellcolor{oursblue}\best{84.67} \\
    \midrule
    \multirow{3}{*}{\textbf{LooGLE-v2}}
      & \textit{Score}
        & 20.76 & \sbest{23.24} & 22.49 & 22.79 & 21.06 & 22.56
        & \cellcolor{oursblue}\best{23.54} \\
      & \textit{Len}
        & 0 & 1621.56 & 5253.94 & 3090.29 & 1 & 6987.13
        & \cellcolor{oursblue}16 \\
      & \textit{CAP} $\uparrow$
        & 20.76 & 25.69 & 23.44 & 24.39 & \sbest{32.54} & 23.18
        & \cellcolor{oursblue}\best{32.67} \\
    \midrule
    \rowcolor{headerblue!15}
    \multicolumn{2}{l}{\textbf{Avg.\ CAP} $\uparrow$}
      & 36.04
      & \sbest{46.63}
      & 43.99
      & 44.58
      & 36.59
      & 45.96
      & \cellcolor{oursblue}\best{55.67} \\
    \bottomrule
  \end{tabular}%
  }
  \arrayrulecolor{black}
\vspace{-2ex}
\end{table}

Table~\ref{tab:main_results} reports results across the three benchmarks. \textbf{OpticalRAG achieves the highest CAP on all three benchmarks while using only 16 injected tokens}, demonstrating a strong balance between task accuracy and inference-time compression efficiency. On LongBench-v2, it obtains the highest raw score of 35.79, improving the closed-book backbone by 5.17 points while using orders of magnitude fewer tokens than textual retrieval and compression baselines.  On BAMBOO, although RECOMP and LLMLingua-v2 attain higher raw scores by retaining hundreds to thousands of textual tokens, our 16-token method still reaches a competitive 61.00 and the best CAP of 84.67, indicating the most favorable accuracy-efficiency trade-off. On LooGLE-v2, OpticalRAG achieves both the best raw score (23.54) and the best CAP (32.67); xRAG attains a similar CAP only because its single-token budget inflates the multiplier, while its raw score remains close to the closed-book backbone, suggesting limited evidence preservation. Overall, conventional textual retrieval and compression methods rely on large token budgets to maintain accuracy, while extremely aggressive compression can lose task-critical information; by compressing long documents into a small number of query-driven visual tokens, OpticalRAG attains a better balance between long-context understanding and inference efficiency.

\subsection{Ablation Study}
\label{sec:ablation}

\definecolor{headerblue}{RGB}{31,73,125}
\definecolor{oursblue}{RGB}{219,229,241}
\definecolor{accentline}{RGB}{31,73,125}



\begin{figure}[t]
    \centering
    \begin{minipage}[b]{0.32\textwidth}
        \centering
        \includegraphics[width=\linewidth]{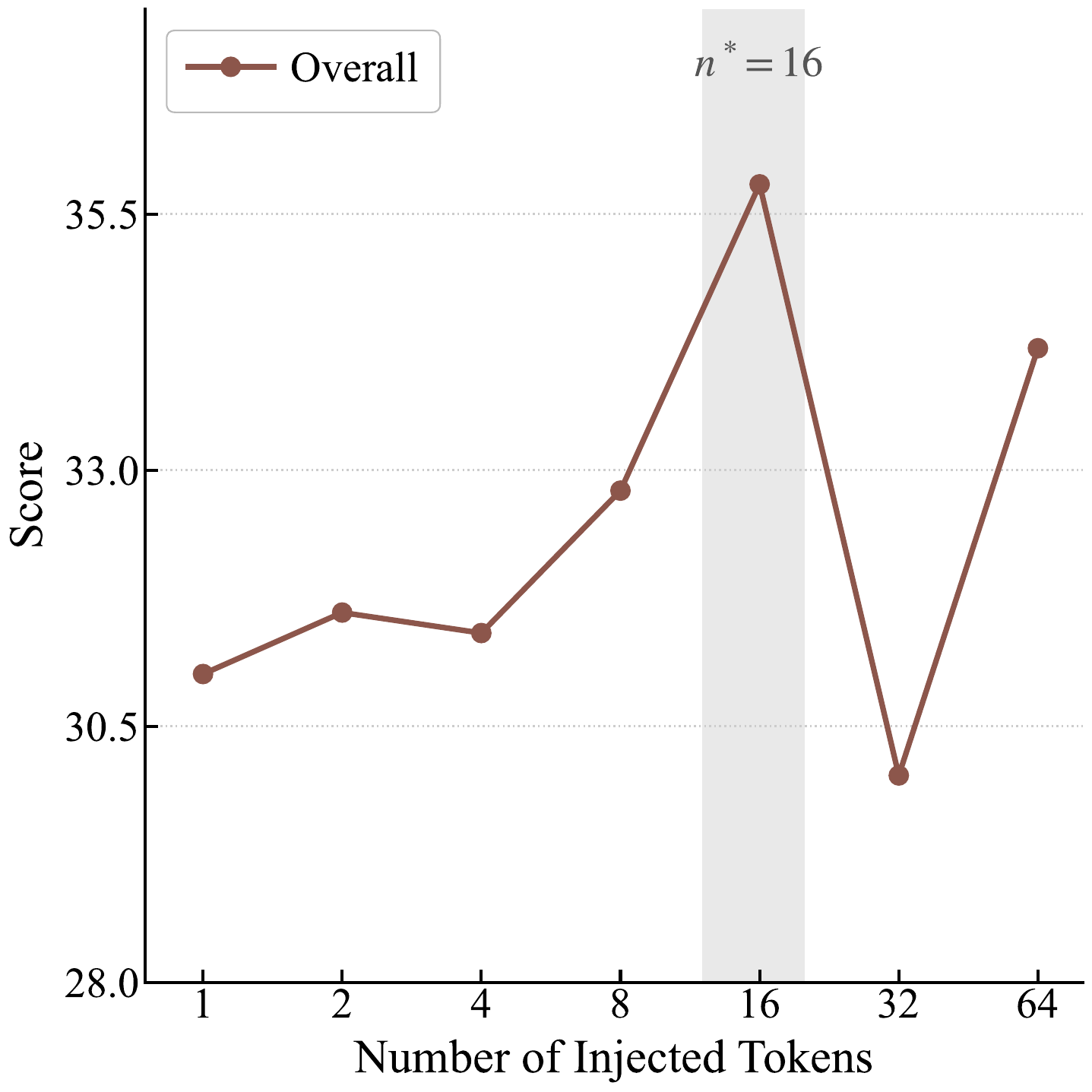}\\[2pt]
        \small (a) Overall
    \end{minipage}
    \hfill
    \begin{minipage}[b]{0.32\textwidth}
        \centering
        \includegraphics[width=\linewidth]{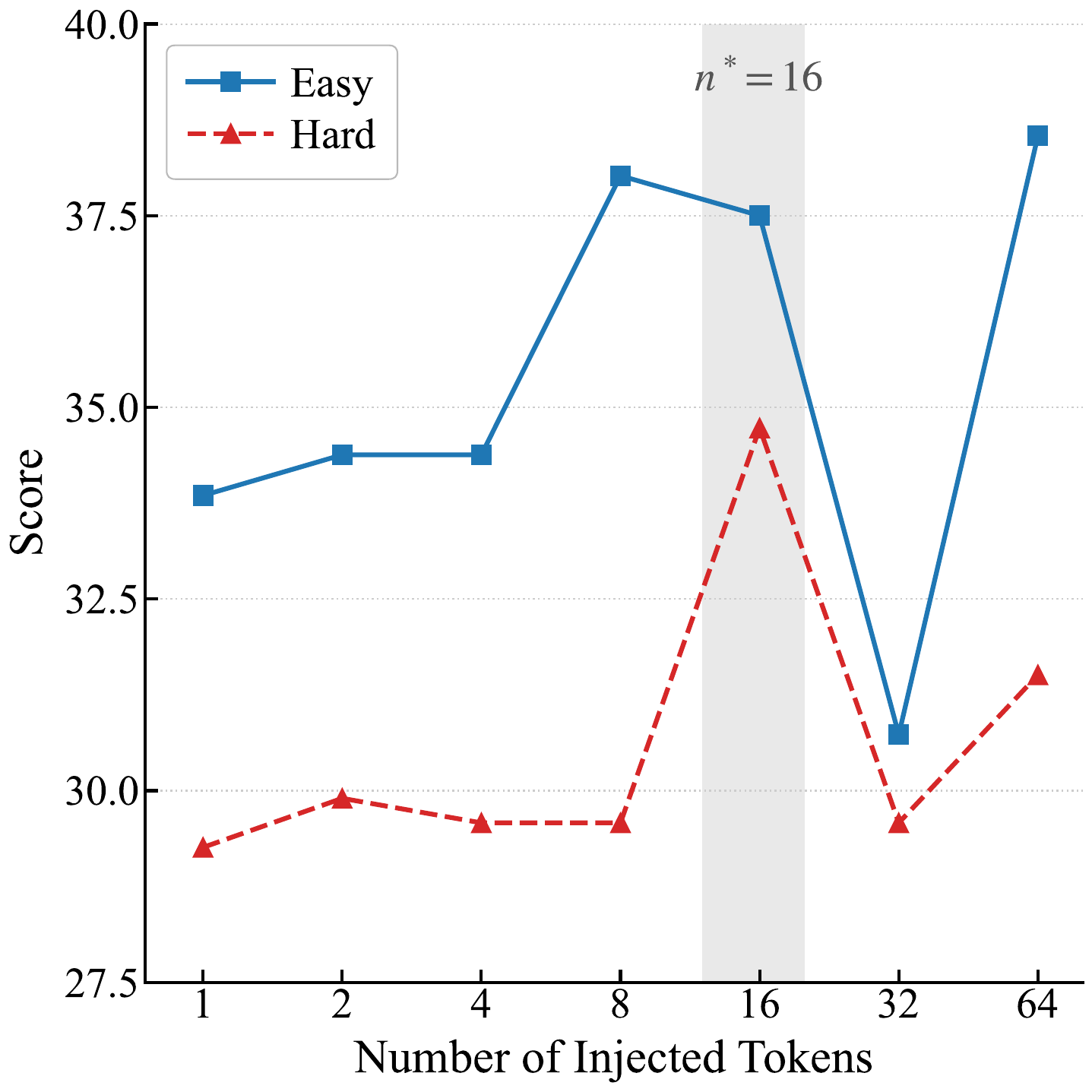}\\[2pt]
        \small (b) Easy \& Hard
    \end{minipage}
    \hfill
    \begin{minipage}[b]{0.32\textwidth}
        \centering
        \includegraphics[width=\linewidth]{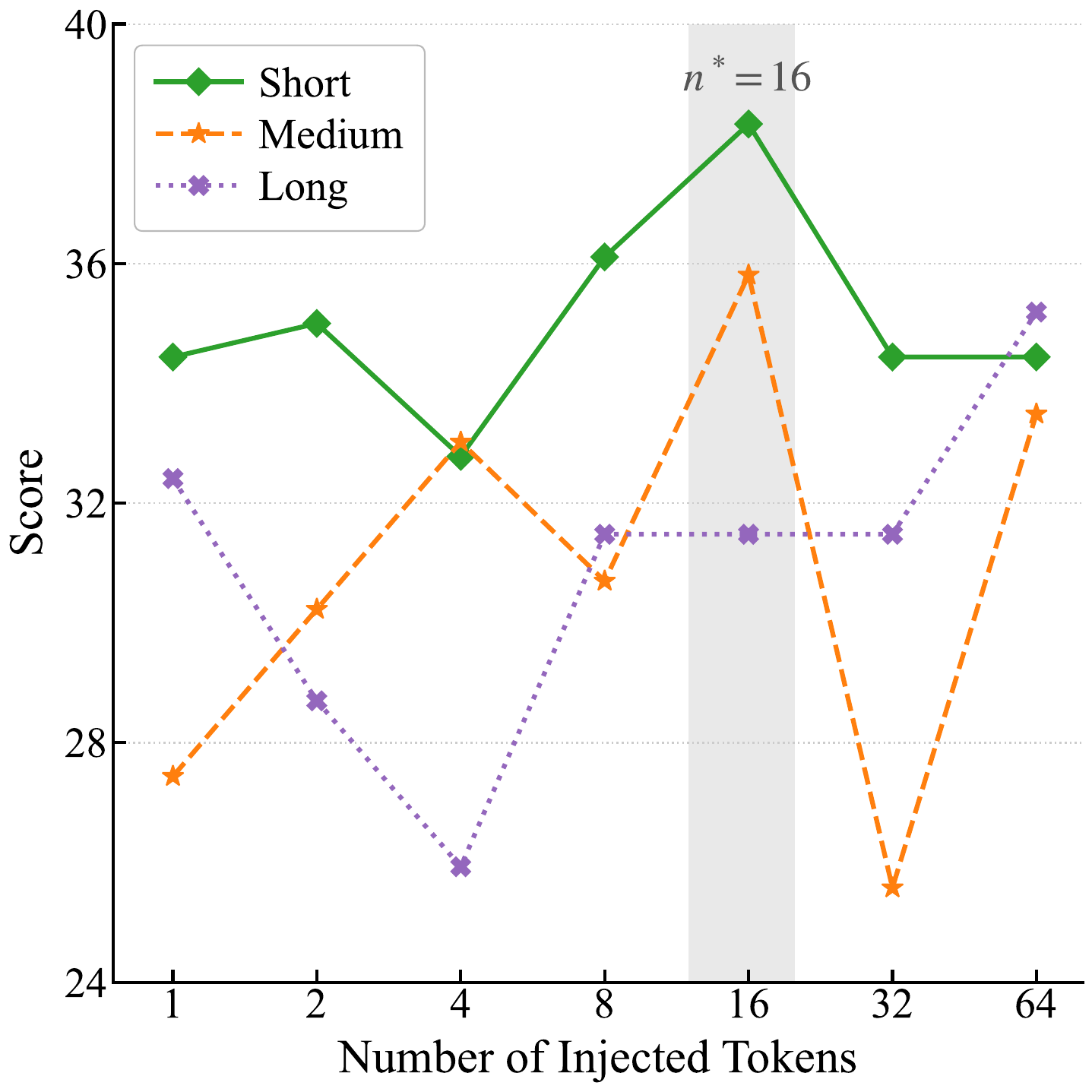}\\[2pt]
        \small (c) Short, Medium \& Long
    \end{minipage}
    \caption{Ablation on the number of injected tokens. }
    \label{fig:token-ablation}
\end{figure}


\begin{figure}[h]
    \centering
    \includegraphics[width=\linewidth]{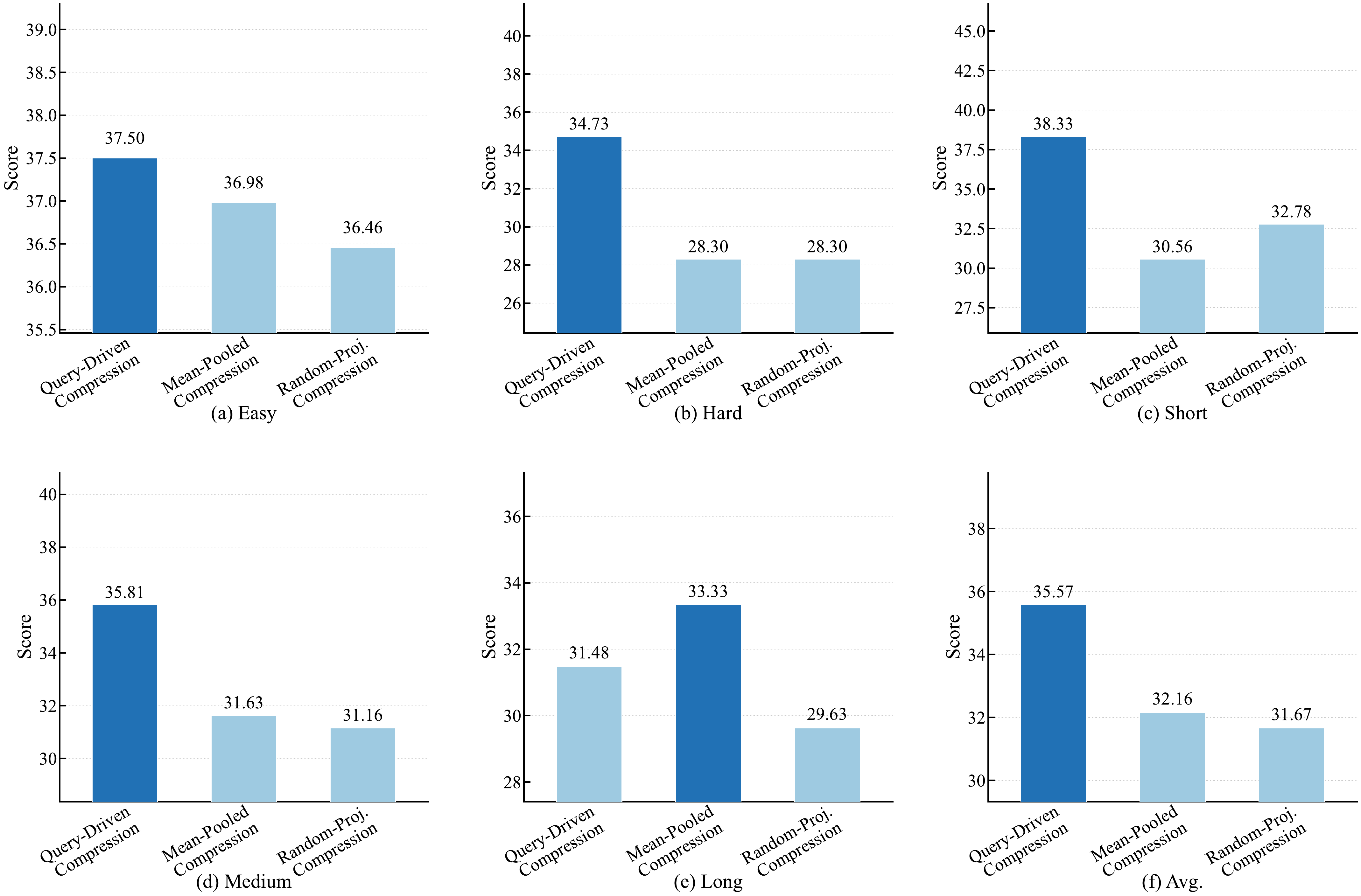}
    \caption{Ablation on query-driven versus query-agnostic compression on LongBench-v2.}
    \label{fig:qdc}
\end{figure}

\paragraph{Number of Injected Tokens.}
As shown in Figure~\ref{fig:token-ablation}, increasing the token budget from $1$ to $16$ generally improves performance, with $N{=}16$ achieving the best overall score of 35.79. Larger budgets do not consistently help: $N{=}32$ leads to a substantial drop and $N{=}64$ only partially recovers, although $N{=}64$ achieves the best score on long examples, suggesting that longer inputs benefit from a slightly larger compressed representation. Overall, $N{=}16$ provides the best trade-off across difficulty levels and input lengths, indicating that the optimal token budget should balance evidence coverage against noise.

\paragraph{Query-Driven Compression.}
We evaluate the importance of query conditioning by comparing Query-Driven Compression with two query-agnostic alternatives under the same compression budget. 
As shown in Figure~\ref{fig:qdc}, Query-Driven Compression achieves a score of 35.57, outperforming Mean-Pooled Compression and Random-Projection Compression, which obtain 32.16 and 31.67, respectively. 
This confirms that query-conditioned compression retains task-relevant evidence more effectively than query-agnostic alternatives, consistent with the theoretical analysis in Section~\ref{sec:theory}.

\section{Conclusion and Limitations}

We presented \textbf{OpticalRAG}, a framework that reimagines context compression for retrieval-augmented generation by operating in the pixel space rather than the text token space. Through \emph{Encoder-Bridged Retrieval}, which projects text queries into the frozen visual feature space through a projection shared with the compressor, without additional contrastive training, and \emph{Query-Driven Distillation}, which condenses retrieved visual tokens into a minimal query-conditioned prefix, OpticalRAG attains the best CAP across LongBench-v2, BAMBOO, and LooGLE-v2 with only 16 injected tokens.

While OpticalRAG uses a compact fixed-size visual prefix, a single token budget may not suit all queries, particularly those requiring evidence from multiple distant document segments. Our evaluation focuses on multiple-choice QA; extending the method to open-ended generation and exploring adaptive prefix budgets remain directions for future work.

\begin{ack}
This research was supported by the National Natural Science Foundation of China (NSFC) under Grants No.~62506366, No.~62272466, U24A20233 and No.~62572268, the Ministry of Science and Technology of China under contract No.~2024YFB2809103, the Fundamental Research Funds for the Central Universities, and the Research Funds of Renmin University of China. It was also supported by Big Data and Responsible Artificial Intelligence for National Governance, Renmin University of China, and by the Public Computing Cloud, Renmin University of China. This work was sponsored by the CIPS-SMP-Zhipu Large Model Fund.
\end{ack}


\bibliographystyle{abbrvnat}
\bibliography{references}

\newpage
\appendix
\section{Extended Theoretical Analysis}
\label{app:theory}

This appendix expands on the theorem in Section~\ref{sec:theory} along three directions: (i) an information-theoretic restatement in terms of mutual information and a sufficient-statistic interpretation, (ii) a robustness result for the approximate low-rank case where Eq.~\eqref{eq:sufficiency} holds only up to a residual, and (iii) discussion of the assumptions and their empirical plausibility for the OpticalRAG pipeline.

\subsection{Information-Theoretic Restatement}
\label{app:info}

Let $A^\star$, $V$, and $Q$ denote the random variables corresponding to the answer, the query-fused visual tokens $\hat{\mathbf{V}}_{\text{fused}}$, and the query-global vector $\mathbf{q}_{\text{global}}$, respectively, and let $T := \mathbf{U}^\top [V_{\sigma(1)}, \ldots, V_{\sigma(r)}]$ denote the statistic in Eq.~\eqref{eq:sufficiency}, which is a function of $(V, Q)$. The sufficiency assumption \eqref{eq:sufficiency} is equivalent to the conditional independence
\begin{equation}
    A^\star \;\perp\!\!\!\perp\; V \;\big|\; \bigl(T,\, Q\bigr),
    \label{eq:cond-indep}
\end{equation}
which by the data-processing inequality implies
\begin{equation}
    I(A^\star ;\, V \mid Q) \;=\; I\!\left(A^\star ;\, T \,\big|\, Q\right).
\end{equation}
Thus $T$ is a \emph{conditional sufficient statistic} of $V$ for $A^\star$ given $Q$. The constructive proof in the main text shows that the cross-attention output $\mathbf{Z}^{(1)}$ encodes this statistic up to an error that vanishes as $\tau \to \infty$. Whenever $T$ can be recovered from $(\mathbf{Z}^{(1)}, Q)$, as in this hard-attention limit, $I(A^\star;\, \mathbf{Z}^{(1)} \mid Q) = I(A^\star;\, V \mid Q)$, i.e., no task-relevant information is lost; for finite $\tau$, the guarantee is the $\varepsilon$-approximation in Eq.~\eqref{eq:lossless}.

\subsection{Approximate Low-Rank Case}
\label{app:approx}

In practice, Eq.~\eqref{eq:sufficiency} holds only approximately. Suppose
\begin{equation}
    p(a^\star \mid V, Q) \;=\; g\!\left(T,\, Q\right) + \varepsilon(V, Q),
\end{equation}
where $\varepsilon(V, Q)$ is a residual with bounded $L^2$ norm $\|\varepsilon\|_2 \leq \delta$ uniformly in $V, Q$. Then by the triangle inequality and the construction in the proof of Theorem~\ref{thm:lossless}, whose attention error is at most $\varepsilon_{\tau} := 2L_g rR(S'-1)e^{-\tau\gamma}$, the output distribution under the cross-attention compressor satisfies
\begin{equation}
    \bigl\|\Phi(\mathbf{Z}^{(1)}, Q) - p(a^\star \mid V, Q)\bigr\|_2 \;\leq\; \delta + \varepsilon_{\tau},
\end{equation}
i.e., the error degrades gracefully with the residual energy. Since $\mathrm{TV} \leq \tfrac{\sqrt{C}}{2}\|\cdot\|_2 = \|\cdot\|_2$ for $C=4$ candidate labels, this suggests that $M=16$ remains adequate when the true conditional rank slightly exceeds $r$: as long as the residual $\delta$ left by the dominant $M$ query-conditioned factors is small, the compression remains near-lossless in total variation.

\subsection{Assumption Plausibility}
\label{app:plausibility}

The low-rank sufficiency assumption is mild for multiple-choice QA over long documents. The four candidate labels lie on a 3-dimensional probability simplex, so the answer distribution itself has at most three degrees of freedom. The query specifies which document fragments are relevant, and the correct answer typically depends on only a handful of facts---names, dates, numerical comparisons, or causal relations. Empirically, this matches the observation that $M=16$ already saturates performance in our ablations: increasing $M$ beyond this point does not improve overall performance (Table~\ref{tab:token_ablation}), consistent with $r \leq 16$ for the LongBench-v2 distribution. A more refined analysis would estimate $r$ directly via the spectrum of the query-conditioned Jacobian of $g$, which we leave to future work.

\subsection{Relation to the Q-Former Bottleneck}
\label{app:qformer}

The construction in the proof generalizes the Q-Former design: the learnable queries $\mathbf{Z}$ act as ``probes'' into the visual token pool, with cross-attention selecting $r$ informative locations. The key new ingredient relative to the original Q-Former is the \emph{query-conditioning} of the bottleneck through $\mathbf{q}_{\text{global}}$, which makes the selected locations $\sigma(j)$ depend on the question. Without this conditioning, a single fixed bottleneck would have to cover the union of all task-relevant evidence across queries, forcing $M$ to scale with the diversity of possible questions rather than the per-query intrinsic rank. Note that adding $\mathbf{q}_{\text{global}}$ to every visual token alone does not provide this: with a linear key projection, it shifts all attention logits of a query by the same constant and leaves the softmax weights unchanged. The query dependence of the selection therefore comes from conditioning the compression queries, $\mathbf{Z}^{(0)} = \mathbf{Z} + \mathbf{1}_M \mathbf{q}_{\text{global}}^\top$. This explains why query conditioning in OpticalRAG's main pipeline is not merely a heuristic but a structural requirement for the bound $M \geq r$ to apply per query rather than globally.

\section{Evaluation Datasets and Protocol}
\label{app:evaluation_dataset}

\subsection{Dataset Details}
\label{app:datasets}

This appendix provides details of the datasets used in our baseline reproduction and evaluation. 
We conduct experiments on three long-context multiple-choice benchmarks: LongBench-v2 \citep{DBLP:conf/acl/BaiTZ0WLCX0D0L25}, the BAMBOO choice subset \citep{DBLP:conf/coling/DongTLZW24} , and the LooGLE-v2 \citep{DBLP:conf/nips/HeWLLZ25} choice subset.

\subsubsection{Dataset Splits}

We use the following dataset splits and subsets in our experiments:
\begin{itemize}
    \item \textbf{LongBench-v2}: the official train split.
    \item \textbf{BAMBOO}: the merged choice-question subset.
    \item \textbf{LooGLE-v2}: the merged choice-question subset.
\end{itemize}

\subsubsection{Dataset Statistics}

Table~\ref{tab:dataset_overview} summarizes the scale and file format of each dataset used in our experiments.

\begin{table}[h]
\centering
\begin{tabular}{lccc}
\hline
Dataset & Format & Samples & Notes \\
\hline
LongBench-v2 train & HuggingFace Arrow & 503 & 6 domains / 16 sub-domains \\
BAMBOO choice subset & JSONL & 400 & 4 tasks, 100 samples each \\
LooGLE-v2 choice subset & JSONL & 1334 & 6 tasks with mixed label spaces \\
\hline
\end{tabular}
\caption{Statistics of the datasets used in our experiments.}
\label{tab:dataset_overview}
\end{table}

\subsubsection{LongBench-v2}

We use the train split of LongBench-v2, which contains 503 samples. Each sample is a four-way multiple-choice question paired with a long context and a single gold answer label from \texttt{A/B/C/D}. The dataset covers 6 coarse domains and 16 sub-domains.

The local data fields include:
\texttt{\_id}, \texttt{domain}, \texttt{sub\_domain}, \texttt{difficulty}, \texttt{length}, \texttt{question}, 
\texttt{choice\_A}, \texttt{choice\_B}, \texttt{choice\_C}, \texttt{choice\_D}, \texttt{answer}, and \texttt{context}.

\subsubsection{BAMBOO Choice Subset}

We evaluate on the choice-question subset of BAMBOO. According to the dataset README, this subset contains four tasks:
\begin{itemize}
    \item \texttt{meetingqa\_4k}: 100 samples;
    \item \texttt{meetingqa\_16k}: 100 samples;
    \item \texttt{paperqa\_4k}: 100 samples;
    \item \texttt{paperqa\_16k}: 100 samples.
\end{itemize}
This results in 400 samples in total.

The merged evaluation file preserves the original option set of each sample. Therefore, the number of options is not fixed across samples. In our local copy, most examples are four-way multiple-choice questions, while some samples contain 1, 2, 3, 5, 6, or 7 options.

Representative fields include:
\texttt{id}, \texttt{benchmark}, \texttt{task}, \texttt{question}, \texttt{context}, \texttt{options}, \texttt{option\_labels}, and \texttt{answer\_label}.

\subsubsection{LooGLE-v2 Choice Subset}

We also evaluate on the choice-question subset of LooGLE-v2. According to the dataset README, this subset contains six tasks:
\begin{itemize}
    \item Call Graph Analysis: 376 samples;
    \item Environment Understanding: 176 samples;
    \item Legal Article Extraction: 186 samples;
    \item Legal Case Retrieval: 267 samples;
    \item Rule Understanding: 165 samples;
    \item User Behavior Analysis: 164 samples.
\end{itemize}
This results in 1334 samples in total.

Unlike standard four-way multiple-choice benchmarks, LooGLE-v2 contains mixed answer-label spaces. While many samples use labels such as \texttt{A/B/C/D}, some legal tasks use task-specific labels such as \texttt{CASE\_x} or \texttt{LAW\_x}. Therefore, the option cardinality varies across samples, including 2, 4, 5, 6, 7, and larger option sets.

Representative fields include:
\texttt{sample\_uid}, \texttt{benchmark}, \texttt{task}, \texttt{task\_slug}, \texttt{question}, \texttt{context}, \texttt{options}, \texttt{option\_labels}, and \texttt{answer\_label}.

\subsection{Evaluation Protocol}
\label{app:evaluation_protocol}

We evaluate each method along two axes: task performance and context efficiency.
For LongBench-v2, we use the official \textit{Overall} metric defined by LongBench-v2 as the reported task score.
For the remaining datasets, the reported score is accuracy.
The model is required to predict the option label string, and a prediction is considered correct only when the predicted label exactly matches the ground-truth label.

To measure context efficiency, we log \texttt{lm\_context\_tokens}, which denotes the number of context-only tokens fed into the final language model.
This count excludes the question and candidate options, and therefore reflects the actual document context budget used by each method.
This allows us to compare methods not only by their raw task scores, but also by the amount of context they require at inference time.

\paragraph{Compression-aware score.}
Raw task score alone does not distinguish between methods that achieve strong performance by using a large amount of context and methods that achieve comparable performance with a much smaller token budget.
To jointly evaluate task performance and context efficiency, we additionally report \(\mathrm{CAP}\), a compression-aware metric used as a complementary measure to the raw task score.

For each benchmark, let \(S\) denote the benchmark-specific task score, and let \(B\) denote the context-token budget measured by \texttt{lm\_context\_tokens}.
Let \(S_0\) be the closed-book backbone score on the same benchmark, and let \(B_0=12\text{k}\) be the reference backbone context budget, corresponding to the truncation limit used by Qwen2.5-7B-Instruct.
We first define the compression multiplier as:
\begin{equation}
  m(B) = \left(\frac{B_0 + 1}{B + 1}\right)^{\delta},
  \qquad \delta = 0.05.
  \label{eq:cap_multiplier}
\end{equation}
The \(+1\) term avoids numerical instability when \(B\) is very small.
The exponent \(\delta\) controls the strength of the compression reward.
We set \(\delta=0.05\) to make the compression reward conservative: a \(10\times\) reduction in context budget gives a multiplier of approximately \(1.122\), while a \(100\times\) reduction gives a multiplier of approximately \(1.259\).

The final \(\mathrm{CAP}\) score is defined as:
\begin{equation}
  \mathrm{CAP}(S, B) =
  \begin{cases}
    S \cdot m(B), & \text{if } S \ge S_0, \\[0.6ex]
    S,            & \text{if } S < S_0.
  \end{cases}
  \label{eq:cap_star_app}
\end{equation}

The lower branch implements a score floor.
When a method performs worse than the closed-book backbone, it receives no compression credit, even if it uses fewer tokens.
This prevents a method from being rewarded for reducing context length while discarding task-critical information.
In other words, compression is credited only when the method preserves at least the closed-book level of task performance.

Above this floor, \(\mathrm{CAP}\) rewards both higher task score and lower context-token budget.
When \(B=B_0\), the multiplier \(m(B)\) is approximately \(1\), so \(\mathrm{CAP}\) reduces to the raw task score for methods that use the same budget as the backbone.
As \(B\) decreases, \(m(B)\) increases according to a mild power law, giving efficient methods additional credit while preventing the metric from being dominated by token reduction alone.

\paragraph{Computation protocol.}
We compute \(\mathrm{CAP}\) separately for each benchmark using the corresponding closed-book backbone score \(S_0\).
The same reference budget \(B_0=12\text{k}\) is used across all benchmarks.
The task score \(S\) is kept in its original benchmark scale, so \(\mathrm{CAP}\) is reported in the same score units as the raw benchmark metric, with an additional compression-aware adjustment.
In all result tables, we report the raw score, the context-token budget, and \(\mathrm{CAP}\), so that task performance and compression efficiency can be inspected independently.

\subsection{Existing Assets and Licenses}
\label{app:asset-licenses}

Tables~\ref{tab:asset-data-licenses} and~\ref{tab:asset-model-licenses}
identify the upstream sources and their published license notices for the
training data, evaluation benchmarks, pretrained components, and rendering
font described in this paper. The linked source pages provide the license
texts or associated terms; a repository license does not supersede separate
rights in third-party content.

\begin{table}[ht]
\centering\small
\renewcommand{\arraystretch}{1.15}
\caption{Dataset sources and published license notices. Source-specific
qualifications are explained below.}
\label{tab:asset-data-licenses}
\begin{tabularx}{\linewidth}{@{}p{2.1cm}p{2.9cm}X@{}}
\toprule
Asset & Source / provider & Published license notice \\
\midrule
LongBench-v2 & \href{https://huggingface.co/datasets/zai-org/LongBench-v2}{THUDM / Z.ai} & Apache-2.0 (dataset card). \\
BAMBOO & \href{https://github.com/RUCAIBox/BAMBOO}{RUCAIBox} & MIT (repository release); underlying source content retains its own terms. \\
LooGLE-v2 & \href{https://huggingface.co/datasets/GraphPKU/LooGLE-v2}{GraphPKU} & CC BY-SA 4.0 (dataset card). \\
Wikipedia & \href{https://github.com/facebookresearch/atlas}{Atlas / Wikimedia} & CC BY-SA 3.0; \texttt{enwiki-dec2021} release. \\
C4 & \href{https://huggingface.co/datasets/allenai/c4}{Google / Ai2 distribution} & ODC-BY for the Ai2 distribution, together with Common Crawl terms. \\
SQuAD v2 & \href{https://rajpurkar.github.io/SQuAD-explorer/}{Stanford SQuAD} & CC BY-SA 4.0. \\
DROP & \href{https://huggingface.co/datasets/ucinlp/drop}{UCI NLP / Hugging Face} & CC BY-SA 4.0; \texttt{drop} / \texttt{ucinlp/drop}. \\
CoQA & \href{https://stanfordnlp.github.io/coqa/}{Stanford CoQA} & Source-dependent: CC BY-SA 4.0, MSR-LA, RACE terms, and Apache terms. \\
\bottomrule
\end{tabularx}
\end{table}

\paragraph{Source-specific data terms.}
The Wikipedia passages use the Atlas \texttt{enwiki-dec2021} release,
whose \href{https://github.com/facebookresearch/atlas#data-license}{data license}
is CC BY-SA 3.0. DROP is obtained through the Hugging Face
\texttt{drop} / \texttt{ucinlp/drop} distribution, whose dataset card
specifies CC BY-SA 4.0.
The C4 distribution's ODC-BY notice additionally refers to
\href{https://commoncrawl.org/terms-of-use}{Common Crawl's terms of use};
those terms preserve separate conditions on crawled content.
CoQA's official page assigns CC BY-SA 4.0 to literature and Wikipedia
passages, MSR-LA to MCTest children's stories, RACE-specific terms to exam
passages, and Apache terms to CNN news passages. CoQA is therefore not
represented as a uniformly licensed corpus.

\begin{table}[ht]
\centering\small
\renewcommand{\arraystretch}{1.15}
\caption{Pretrained components and rendering font: upstream license notices.}
\label{tab:asset-model-licenses}
\begin{tabularx}{\linewidth}{@{}p{3.2cm}p{2.3cm}X@{}}
\toprule
Asset & Source / provider & Published license notice \\
\midrule
DeepSeek-OCR & \href{https://huggingface.co/deepseek-ai/DeepSeek-OCR}{DeepSeek} & MIT (model card and \href{https://github.com/deepseek-ai/DeepSeek-OCR/blob/main/LICENSE}{repository}). \\
Qwen2.5-7B-Instruct & \href{https://huggingface.co/Qwen/Qwen2.5-7B-Instruct}{Qwen / Alibaba} & Apache-2.0. \\
SAM implementation & \href{https://github.com/facebookresearch/segment-anything/blob/main/LICENSE}{Meta} & Apache-2.0. \\
CLIP implementation & \href{https://github.com/openai/CLIP/blob/main/LICENSE}{OpenAI} & MIT. \\
DejaVu Sans Mono & \href{https://dejavu-fonts.github.io/License.html}{DejaVu / Bitstream} & Bitstream Vera font terms; DejaVu modifications are public domain, with additional notices for imported glyphs. \\
\bottomrule
\end{tabularx}
\end{table}

The SAM/CLIP visual weights used here are loaded from the DeepSeek-OCR
checkpoint; their upstream implementation notices are listed separately
from the checkpoint release license. Applicable attribution, license-notice,
share-alike, and source-specific conditions remain attached to the
corresponding assets and are not replaced by the OpticalRAG release license.

\section{Baseline Details}
\label{app:baseline_adaptations}

This appendix provides additional details on how we adapt and reproduce the baseline methods under a unified evaluation setting. 
For fair comparison, all reproduced methods use \texttt{Qwen2.5-7B-Instruct} as the final answer model. 
Our adaptation follows a minimal-modification principle: we preserve the core mechanism of each baseline as much as possible, and only modify components required for compatibility with the unified backbone, input format, and multiple-choice evaluation protocol.

\subsection{Unified Backbone and Evaluation Interface}

\paragraph{Backbone.}
All baselines use \texttt{Qwen2.5-7B-Instruct} as the final decoder for answer generation. 
This avoids performance differences caused by using different downstream language models and allows us to focus on the effectiveness of each compression or retrieval mechanism.

\paragraph{Evaluation interface.}
All methods are evaluated under the same multiple-choice setting. 
After method-specific context compression, selection, or representation injection, we construct a unified choice prompt and require the model to output one label from the valid option set of each dataset.

\paragraph{Adaptation principle.}
We do not replace baseline-specific mechanisms with a generic retrieval or summarization pipeline. 
Instead, each method keeps its original design pattern, such as direct prompt compression, self-information-based filtering, retrieval-embedding injection, or compress-then-generate inference. 
Only the final decoding interface and necessary dimensional or formatting components are adapted.

\subsection{Baseline Methods}

\paragraph{Backbone.}
For the Backbone baseline, we directly use \texttt{Qwen2.5-7B-Instruct} \citep{DBLP:journals/corr/abs-2412-15115} for multiple-choice answer generation. 
No additional training, retrieval, compression, or context injection is performed. 
The model is prompted only with the question and candidate options, without access to the long input context. 
This baseline measures the backbone model's parametric knowledge, choice bias, and direct multiple-choice answering ability in the absence of external evidence.

\paragraph{StandardRAG.}
For StandardRAG, we reproduce the naive RAG setting following the original Meta-RAG execution path \citep{DBLP:journals/corr/abs-2510-24003}. 
The long input context is first split into overlapping character chunks, which are ranked by a BM25-style scorer using the question and, when applicable, the option text. 
We then select the top-\(k\) chunks as the retrieved context and feed them, together with the question and candidate options, into \texttt{Qwen2.5-7B-Instruct} for final label generation. 
No additional compression module, learned retriever, or representation injection mechanism is introduced. 
Unless otherwise specified, we use \(k=5\), a chunk size of 1200 characters, and an overlap of 200 characters.

\paragraph{LLMLingua-2.}
For LLMLingua-2 \citep{DBLP:conf/acl/PanWJXLZLR0LZQ024}, we preserve its original direct prompt compression pipeline implemented by
\texttt{PromptCompressor(..., use\_llmlingua2=True)}. 
The full input context is compressed by LLMLingua-2 without adding an extra retrieval or top-$k$ preselection stage. 
The resulting compressed context is then inserted into the unified multiple-choice prompt and decoded by \texttt{Qwen2.5-7B-Instruct}.

\paragraph{xRAG.}
For xRAG \citep{DBLP:conf/nips/0002W00CWZ024}, we preserve its retrieval-embedding injection paradigm, where retrieved document representations are projected into the language model hidden space and used to replace the embeddings of special \texttt{<xRAG>} placeholders before decoding. 
To make this mechanism compatible with our unified backbone, we implement a Qwen-compatible xRAG wrapper and train only the projector that aligns the retriever embedding dimension with the hidden dimension of \texttt{Qwen2.5-7B-Instruct}. 
The final answer is generated by the same Qwen decoder used by all other methods.

\paragraph{Selective Context.}
For Selective Context \citep{DBLP:conf/emnlp/0001DGL23}, we preserve its original lexical filtering pipeline: the input is segmented into lexical units, self-information scores are estimated, and less informative units are filtered according to the specified \texttt{reduce\_ratio} and \texttt{reduce\_level}. 
Since the original implementation relies on GPT-2 for self-information estimation, we replace GPT-2 with \texttt{Qwen2.5-7B-Instruct} to maintain backbone consistency. 
The method is still treated as full-context compression rather than retrieval-based top-$k$ selection.

\paragraph{RECOMP.}
For RECOMP \citep{DBLP:conf/iclr/XuSC24}, we preserve its compress-then-generate inference structure. 
Because our evaluation setting contains long single-document contexts rather than standard multi-document retrieval outputs, we adapt the compression stage by splitting the context into sentence-level candidates, scoring them by query relevance, and retaining the top-ranked sentences according to a fixed compression ratio. 
The compressed context is then passed to \texttt{Qwen2.5-7B-Instruct} for multiple-choice answer generation.

%

\definecolor{optblue}{RGB}{26,86,155}
\definecolor{optlightblue}{RGB}{235,242,252}
\definecolor{optmidblue}{RGB}{174,206,237}
\definecolor{optgold}{RGB}{160,118,0}
\definecolor{optgoldlight}{RGB}{255,250,220}
\definecolor{optgreen}{RGB}{0,120,80}
\definecolor{optgreenlight}{RGB}{218,244,232}
\definecolor{optdarkgray}{RGB}{75,78,90}
\definecolor{frozencolor}{RGB}{80,130,195}
\definecolor{traincolor}{RGB}{0,145,85}
\definecolor{rowA}{RGB}{248,250,254}
\definecolor{rowB}{RGB}{255,255,255}
\definecolor{hdrfill}{RGB}{26,86,155}
\definecolor{secbar}{RGB}{26,86,155}


\newtcolorbox{infobox}[2][]{%
  enhanced, breakable,
  colback=optlightblue, colframe=optblue,
  coltitle=white, fonttitle=\bfseries\small,
  title={#2}, arc=3pt, boxrule=0.65pt,
  left=7pt, right=7pt, top=4pt, bottom=5pt, #1}

\newtcolorbox{goldbox}[2][]{%
  enhanced, breakable,
  colback=optgoldlight, colframe=optgold,
  coltitle=white, fonttitle=\bfseries\small,
  title={#2}, arc=3pt, boxrule=0.65pt,
  left=7pt, right=7pt, top=4pt, bottom=5pt, #1}

\newtcolorbox{greenbox}[2][]{%
  enhanced, breakable,
  colback=optgreenlight, colframe=optgreen,
  coltitle=white, fonttitle=\bfseries\small,
  title={#2}, arc=3pt, boxrule=0.65pt,
  left=7pt, right=7pt, top=4pt, bottom=5pt, #1}

\newcommand{\frozenbadge}{%
  \tikz[baseline=-1pt]\node[inner sep=2pt,rounded corners=2pt,
    fill=frozencolor!18,draw=frozencolor!55,
    text=frozencolor!80!black,font=\bfseries\scriptsize]{FROZEN};}
\newcommand{\trainbadge}{%
  \tikz[baseline=-1pt]\node[inner sep=2pt,rounded corners=2pt,
    fill=traincolor!18,draw=traincolor!55,
    text=traincolor!80!black,font=\bfseries\scriptsize]{TRAINABLE};}

\newcommand{\secruledeco}{%
  \noindent\textcolor{secbar}{\rule{2.2pt}{9pt}}\hspace{4pt}}

\section{Implementation Details}
\label{app:impl}

\vspace{1pt}
\noindent{\small\color{optdarkgray}%
This appendix describes the implementation and fixed LongBench-v2 inference
configuration in the accompanying code package. Library defaults and optional
variants are distinguished from this configuration.
We organise the material as follows:
\S\ref{app:impl:arch} covers the full architecture and parameter budget;
\S\ref{app:impl:render} details the text-to-image rendering pipeline;
\S\ref{app:impl:encoder} describes the \textsc{DeepEncoder};
\S\ref{app:impl:qdc} specifies the \textsc{QueryDrivenCompressor} in full;
\S\ref{app:impl:scoring} enumerates all retrieval scoring functions;
\S\ref{app:impl:sampling} describes context-window sampling;
\S\ref{app:impl:training} lists the inference configuration and training procedure;
\S\ref{app:impl:eval} documents the answer-extraction protocol.%
}

\bigskip

\subsection{Full Model Architecture and Parameter Budget}
\label{app:impl:arch}

\textsc{OpticalRAG} comprises four modules with clearly separated roles and
training regimes.
Table~\ref{tab:app:arch} summarises each module, its sub-components,
allocated parameter count where verified, and training status.

\begin{table}[h]
\centering
\setlength{\tabcolsep}{4.5pt}
\renewcommand{\arraystretch}{1.38}
\caption{%
  \textbf{Model architecture and parameter budget.}
  Compressor and bridge counts are computed from the instantiated modules.
  Their allocated parameters total $30{,}716{,}928$ ($\approx30.72$ M).%
}
\label{tab:app:arch}
\resizebox{\linewidth}{!}{%
\begin{tabular}{%
  >{\bfseries\small}p{2.7cm}
  p{5.4cm}
  r
  c
  p{5.6cm}
}
\toprule
\rowcolor{hdrfill}
\textcolor{white}{\textbf{Module}} &
\textcolor{white}{\textbf{Sub-components}} &
\textcolor{white}{\textbf{Params}} &
\textcolor{white}{\textbf{Status}} &
\textcolor{white}{\textbf{Role}} \\
\midrule
\rowcolor{rowA}
DeepEncoder &
SAM-ViT-B $+$ CLIP-L; loaded from DeepSeek-OCR checkpoint via \texttt{safetensors} &
--- &
\frozenbadge &
Maps $640{\times}640$ page images to $T{=}100$ visual tokens of dim $d_\text{vis}{=}1024$. \\
\rowcolor{rowB}
Qwen2.5-7B-Instruct &
Hidden dim 3584; loaded from the pinned Qwen checkpoint &
--- &
\frozenbadge &
Provides query hidden states; scores final answer choices via next-token logits. \\
\rowcolor{rowA}
\makecell[tl]{QueryDriven\\Compressor} &
\makecell[tl]{%
  $\mathbf{W}_q\!\in\!\mathbb{R}^{1024\times3584}$,\;
  $\text{DSConv}_{\downarrow2}$,\\
  $2{\times}\text{MHA}$ (8 heads, dim 1024),\\
  learned queries $\mathbf{Z}\!\in\!\mathbb{R}^{16\times1024}$,\; LN} &
$14{,}187{,}520$ &
\trainbadge &
Projects query into visual space; downsamples and cross-attends visual tokens
into $M{=}16$ compressed tokens conditioned on the query. \\
\rowcolor{rowB}
\makecell[tl]{VisualToLLM\\Bridge} &
\makecell[tl]{%
  Linear$(1024{\to}3584)$,\; GELU,\\
  Dropout$(0.1)$,\\
  Linear$(3584{\to}3584)$,\; LN} &
$16{,}529{,}408$ &
\trainbadge &
Projects 16 compressed visual tokens into the LLM's embedding space;
prepended as a soft prefix to the query. \\
\bottomrule
\end{tabular}%
}
\end{table}

\begin{infobox}{Parameter accounting}
The compressor contains $14{,}187{,}520$ parameters and the bridge contains
$16{,}529{,}408$; the bridge accounts for $53.81\%$ of their combined count.
Each PyTorch multi-head attention layer includes its own input and output
projections, in addition to the explicit $\mathbf{W}_K$ and $\mathbf{W}_V$.
The compressor count includes the 1,024-parameter attention-pooling scorer,
which is allocated in every configuration but unused with max pooling.
Frozen-backbone counts and a training-to-total ratio are not reported here
pending a count from the exact model weights.
\end{infobox}

\bigskip

\subsection{Text-to-Image Rendering Pipeline}
\label{app:impl:render}

Each text window is rendered to a sequence of page images by a deterministic
typesetting engine (\texttt{render.py}) using fixed density parameters in the released inference entry point.
The library also exposes an optional piecewise policy
(\texttt{render\_density\_policy.py}).

\paragraph{Font and canvas.}
Pages use \textsc{DejaVuSansMono}, selected via
\texttt{V23\_RENDER\_FONT\_PATH} or one of two standard Linux font paths.
A missing font raises an error; there is no default-font fallback.
The reproduction entry point verifies the font SHA256 recorded in
\texttt{configs/longbench.json}.
Every page is rendered at a fixed resolution of $H{\times}H = 640{\times}640$
pixels with a white background and black text.

\paragraph{Fixed density and estimated capacity.}
The released configuration uses $s_f=15$ pixels, $s_m=20$ pixels, and
$s_l=1.3$. The renderer computes
\[
 h_{\rm line}=\lfloor15\cdot1.3\rfloor=19,\quad
 w_{\rm char}=\max(6,\lfloor0.6\cdot15\rfloor)=9,\quad
 H-2s_m=600.
\]
Thus the nominal capacity is $\lfloor600/9\rfloor=66$ characters per line
and $\lfloor600/19\rfloor=31$ lines per page, or 2,046 characters per page.
This is a wrapping estimate, not an exact count of rendered characters:
the implementation strips outer whitespace, replaces tabs with spaces,
splits on spaces, wraps words, and hard-wraps overlong words.
The library defaults of $(16,24,1.45)$ instead give an estimated width of
9 pixels, 65 characters per line, and 25 lines per page (1,625 nominal
characters); these defaults are overridden by the released entry point.

\paragraph{Length-adaptive density policy.}
The optional policy below increases font size, margins, or line spacing
as context length increases, thereby decreasing nominal page density.
It is not used by the fixed LongBench-v2 reproduction configuration.
Table~\ref{tab:app:density} shows the three-tier schedule activated when the
environment variable \texttt{V23\_RENDER\_POLICY\_MODE=length} is set.

\begin{table}[h]
\centering
\setlength{\tabcolsep}{8pt}
\renewcommand{\arraystretch}{1.25}
\caption{\textbf{Length-adaptive rendering density tiers.}}
\label{tab:app:density}
\begin{tabular}{clccc}
\toprule
\rowcolor{hdrfill}
\textcolor{white}{\textbf{Tier}} &
\textcolor{white}{\textbf{Context length threshold}} &
\textcolor{white}{$s_f$ (px)} &
\textcolor{white}{$s_m$ (px)} &
\textcolor{white}{$s_l$} \\
\midrule
\rowcolor{rowA}
1 & $\leq 300{,}000$ chars  & 15 & 20 & 1.30 \\
\rowcolor{rowB}
2 & $\leq 900{,}000$ chars  & 15 & 22 & 1.35 \\
\rowcolor{rowA}
3 & $> 900{,}000$ chars     & 16 & 24 & 1.45 \\
\bottomrule
\end{tabular}
\end{table}

\paragraph{Bytepack mode.}
An alternative mode (\texttt{full\_visual\_render\_mode=bytepack}) encodes raw
UTF-8 bytes directly into pixel intensity values, packing up to
$640^2 = 409{,}600$ bytes per grayscale page, later converted to RGB.
This is the capacity of this byte-to-pixel scheme, not a general
information-theoretic upper bound, and it discards all
typographic structure that the visual encoder was originally trained on.
It is an optional implementation and is not selected by the released
LongBench-v2 entry point.

\paragraph{Image pre-processing.}
Before entering \textsc{DeepEncoder}, each page image is:
(1) resized to $640{\times}640$ via bicubic interpolation;
(2) converted to \texttt{float32} in $[0,1]$;
(3) normalised as $x \leftarrow (x - 0.5) / 0.5$, yielding values in $[-1, 1]$.
Images are batched in groups of 4 to balance GPU memory usage with throughput.

\bigskip

\subsection{DeepEncoder Architecture}
\label{app:impl:encoder}

The visual encoder (\texttt{encoder.py}) loads two sub-networks from the
DeepSeek-OCR \texttt{safetensors} checkpoint: a \textbf{SAM-ViT-B} backbone
(weights prefixed \texttt{model.sam\_model.*}) and a \textbf{CLIP-L} visual
tower (weights prefixed \texttt{model.vision\_model.*}).
The complete forward pass is:

\begin{align}
  \mathbf{F}_\text{sam}
    &= \textsc{SAM}\!\left(\mathbf{I}\right)
     \in \mathbb{R}^{B\times C\times H'\times W'},\\
  \hat{\mathbf{F}}_\text{sam}
    &= \mathbf{F}_\text{sam}.\texttt{flatten}(2).\texttt{permute}(0,2,1)
     \in \mathbb{R}^{B\times T\times C},
  \label{eq:app:sam}\\[2pt]
  \mathbf{C}_\text{clip}
    &= \textsc{CLIP}\!\left(\mathbf{I},\,\mathbf{F}_\text{sam}\right)
     \in \mathbb{R}^{B\times (T+1)\times 1024},
  \label{eq:app:clip}\\[2pt]
  \mathbf{e}_\text{idx}
    &= \ell_2\text{-norm}\!\left(\mathbf{C}_\text{clip}[:,0,:]\right)
     \in \mathbb{R}^{B\times 1024},
  \label{eq:app:idx}\\[2pt]
  \mathbf{V}
    &= \hat{\mathbf{F}}_\text{sam} + \mathbf{C}_\text{clip}[:,1:,:]
     \in \mathbb{R}^{B\times T\times 1024},\quad T=100.
  \label{eq:app:vis}
\end{align}

\noindent$\mathbf{e}_\text{idx}$ (Eq.~\ref{eq:app:idx}) is the L2-normalised CLS
token returned by the encoder but discarded by the released cache builder.
The index instead stores each fused patch token in $\mathbf{V}$ individually,
without L2 normalisation in dot mode, together with its original position.
A separate FAISS \texttt{IndexIDMap2(IndexFlatIP)} is built for each sample
context; retrieval is restricted to the current context.
$\mathbf{V}$ (Eq.~\ref{eq:app:vis}) is the \emph{additive fusion} of SAM
spatial tokens with CLIP patch tokens, forming the 100-token compressed visual
representation passed to the compressor.
All encoder parameters remain frozen throughout training.

\bigskip

\subsection{QueryDrivenCompressor: Detailed Forward Pass}
\label{app:impl:qdc}

\begin{greenbox}{Shared projection $\mathbf{W}_q$}
The weight matrix $\mathbf{W}_q\in\mathbb{R}^{1024\times3584}$
(\texttt{proj\_q}) is reused by retrieval and compression.
Both stages use the same projected query states; compared with allocating
a second projection of the same shape, this saves $3{,}670{,}016$ parameters.
\end{greenbox}

\vspace{4pt}
\noindent Algorithm~\ref{alg:app:qdc} provides the complete forward pass of
\texttt{QueryDrivenCompressor} with explicit tensor shapes.

\begin{algorithm}[h]
\small
\caption{\textbf{QueryDrivenCompressor} forward pass (from \texttt{query\_driven.py})}
\label{alg:app:qdc}
\begin{algorithmic}[1]
\Require Visual tokens
  $\mathbf{V}\!\in\!\mathbb{R}^{B\times N\times 1024}$,
  query hidden states
  $\mathbf{H}_q\!\in\!\mathbb{R}^{B\times n_q\times 3584}$
\Ensure Compressed tokens
  $\hat{\mathbf{Z}}\!\in\!\mathbb{R}^{B\times M\times 1024}$
\State
  $\tilde{\mathbf{H}}_q \leftarrow \mathbf{H}_q \mathbf{W}_q^\top$
  \hfill\textcolor{optdarkgray}{\small\# LLM$\to$visual space:
    $\mathbb{R}^{B\times n_q\times3584}\!\to\!\mathbb{R}^{B\times n_q\times1024}$}
\State
  $\mathbf{q}_\text{g} \leftarrow \textsc{Pool}(\tilde{\mathbf{H}}_q)$
  \hfill\textcolor{optdarkgray}{\small\# global query vector:
    $\mathbb{R}^{B\times1\times1024}$;\; see Table~\ref{tab:app:qpool}}
\State
  $\mathbf{V} \leftarrow \mathbf{V} + \mathbf{q}_\text{g}$
  \hfill\textcolor{optdarkgray}{\small\# additive query-global bias over all visual tokens}
\State
  $\mathbf{V} \leftarrow \textsc{PreFilter}(\mathbf{V},\mathbf{q}_\text{g})$
  \hfill\textcolor{optdarkgray}{\small\# optional hard or soft query-guided token pruning (§\ref{app:impl:prefilter})}
\State
  $\mathbf{V} \leftarrow \textsc{DSConv}_{\downarrow2}(\mathbf{V})$
  \hfill\textcolor{optdarkgray}{\small\# depthwise conv: $N\!\to\!\lfloor\sqrt{N}/2\rfloor^2$ for square $N$}
\State
  $\mathbf{K} \leftarrow \mathbf{V}\mathbf{W}_K^\top$;\quad
  $\mathbf{Val} \leftarrow \mathbf{V}\mathbf{W}_V^\top$
  \hfill\textcolor{optdarkgray}{\small\#
    $\mathbf{W}_K,\mathbf{W}_V\!\in\!\mathbb{R}^{1024\times1024}$, no bias}
\State
  $\mathbf{Z}^{(0)} \leftarrow \mathbf{Z} + \mathbf{q}_\text{g}$
  \hfill\textcolor{optdarkgray}{\small\# $M{=}16$ learned latents, init std 0.02, query-biased}
\For{$\ell = 1,\ldots,L$}
  \hfill\textcolor{optdarkgray}{\small\# $L{=}2$ cross-attention layers}
  \State
    $\mathbf{Z}^{(\ell)} \leftarrow \mathbf{Z}^{(\ell-1)}
      + \textsc{MHA}^{(\ell)}\!\left(\mathbf{Z}^{(\ell-1)},\,\mathbf{K},\,\mathbf{Val}\right)$
    \hfill\textcolor{optdarkgray}{\small\# residual cross-attention, 8 heads}
\EndFor
\State \Return
  $\hat{\mathbf{Z}} \leftarrow \textsc{LayerNorm}\!\left(\mathbf{Z}^{(L)}\right)$
\end{algorithmic}
\end{algorithm}

\paragraph{Depthwise spatial downsampling (line~5).}
The downsampler is a single \texttt{nn.Conv2d} with
\texttt{in\_channels = out\_channels = 1024},
\texttt{kernel\_size = stride = 2},
\texttt{groups = 1024} (strictly depthwise; no cross-channel mixing), and
\texttt{bias = False}.
When downsampling is enabled, it is applied to square token counts
($\sqrt{N}\in\mathbb{Z}$). The convolution requires $\sqrt{N}\ge2$;
the implementation has no separate guard for smaller square inputs.
Tokens are reshaped $[B,N,d]\to[B,d,\sqrt{N},\sqrt{N}]$, convolved,
and reshaped back to $[B,\lfloor\sqrt{N}/2\rfloor^2,d]$.
In the released configuration, Top-256 retrieval is followed by sorting
the selected tokens by their original positions. The resulting 256 tokens
are reshaped to a $16\times16$ grid and downsampled to 64 tokens.
This grid is a rearrangement of selected tokens, potentially spanning pages
and gaps; it does not reconstruct the original page neighbourhoods.
When fewer than 256 tokens are available, the actual count is used;
non-square counts bypass downsampling, as do all counts when downsampling
is disabled.

\paragraph{Query pooling strategies.}
\label{app:impl:qpool}
Table~\ref{tab:app:qpool} documents all six options for $\textsc{Pool}(\cdot)$
in line~2 of Algorithm~\ref{alg:app:qdc}.

\begin{table}[h]
\centering
\setlength{\tabcolsep}{5pt}
\renewcommand{\arraystretch}{1.32}
\caption{%
  \textbf{Query pooling strategies.}
  Input: $\tilde{\mathbf{H}}_q\!\in\!\mathbb{R}^{B\times n_q\times1024}$.
  Output: $\mathbf{q}_\text{g}\!\in\!\mathbb{R}^{B\times1\times1024}$.
  The \textbf{max} strategy (shaded) is used by the released LongBench-v2 entry point.%
}
\label{tab:app:qpool}
\resizebox{\linewidth}{!}{%
\begin{tabular}{>{\ttfamily\small}l l p{6.5cm}}
\toprule
\rowcolor{hdrfill}
\textcolor{white}{\textbf{Flag value}} &
\textcolor{white}{\textbf{Formula}} &
\textcolor{white}{\textbf{Notes}} \\
\midrule
\rowcolor{rowA}
mean &
$\dfrac{1}{n_q}\displaystyle\sum_{i}\tilde{\mathbf{h}}_{q,i}$ &
Smooth global average; the natural default. \\[6pt]
\rowcolor{optgoldlight}
\textbf{max} &
$\displaystyle\max_{i}\tilde{\mathbf{h}}_{q,i}$ \quad (element-wise) &
\textbf{Released inference configuration.}
Element-wise maximum; pooling does not mask padding positions. \\[4pt]
\rowcolor{rowA}
last &
$\tilde{\mathbf{h}}_{q,n_q}$ &
Last-token summary; analogous to causal LM hidden state. \\
\rowcolor{rowB}
max\_mean &
$\tfrac{1}{2}\!\left(\displaystyle\max_i\tilde{\mathbf{h}}_{q,i}
  + \tfrac{1}{n_q}\sum_i\tilde{\mathbf{h}}_{q,i}\right)$ &
Arithmetic mean of max and mean pooling. \\[4pt]
\rowcolor{rowA}
attn &
$\displaystyle\sum_i w_i\,\tilde{\mathbf{h}}_{q,i}$,\;\;
$w_i = \operatorname{softmax}_i\!\left(\mathbf{a}^\top\tilde{\mathbf{h}}_{q,i}\right)$ &
Learned scalar scoring with $\mathbf{a}\!\in\!\mathbb{R}^{1024}$.
Allocated in all modes, but used only in this mode. \\[4pt]
\rowcolor{rowB}
none &
$\mathbf{0}\in\mathbb{R}^{B\times1\times1024}$ &
Disables query bias in the compressor only; retrieval still uses max-pooled
projected query states. \\
\bottomrule
\end{tabular}%
}
\end{table}

\paragraph{Pre-filtering visual tokens.}
\label{app:impl:prefilter}
Before downsampling, an optional first-pass query-guided filter prunes or
reweights visual tokens based on query relevance
(\texttt{pre\_filter\_topk}, \texttt{pre\_filter\_mode}).

\vspace{2pt}
\begin{itemize}[leftmargin=*, itemsep=3pt, topsep=2pt]
  \item \textbf{Hard top-$K$} (\texttt{hard}): score each token as
    $s_i = \mathbf{v}_i^\top\mathbf{q}_\text{g}$,
    retain the top-$K$ tokens and discard the rest.
    Enabled by setting \texttt{pre\_filter\_topk} $> 0$.
    The released entry point uses \texttt{hard} with $K=0$, disabling this filter.
    Hard selection uses unsorted top-$K$ indices.
  \item \textbf{Soft sigmoid gating} (\texttt{soft}): compute per-token
    relevance weights and rescale:
    \[
      \tilde{\mathbf{v}}_i
        = \mathbf{v}_i \cdot
          \sigma\!\!\left(
            \lambda\cdot\frac{\mathbf{v}_i^\top\mathbf{q}_\text{g}}{\sqrt{d}}
          \right),
      \quad
      \lambda = \max(10^{-6},\texttt{pre\_filter\_soft\_scale})
      \ (\text{default}\ 1.0).
    \]
    This branch applies even when \texttt{pre\_filter\_topk}=0.
    All tokens are retained; low-relevance tokens are multiplicatively
    suppressed rather than discarded.
\end{itemize}

\bigskip

\subsection{Retrieval Scoring Functions}
\label{app:impl:scoring}

The \texttt{QueryVisualScorer} (\texttt{query\_visual\_scoring.py}) defines scores for the online retriever, offline cache scans, and
candidate reranking. Table~\ref{tab:app:scoring} lists its scoring functions.
The main offline FAISS dot path instead returns the raw inner product
$\mathbf{v}^\top\mathbf{q}$, without the $1/\sqrt{d}$ factor; this positive
scaling preserves rankings but changes the numerical scores.
For optional multi-vector modes, the offline index first recalls candidates
with the global query (up to \texttt{retrieval\_rerank\_factor}$\times K$,
default factor 4), then applies the listed function within that candidate set.
The released single-vector dot path retrieves $K$ directly without reranking.

\begin{table}[h]
\centering
\setlength{\tabcolsep}{4pt}
\renewcommand{\arraystretch}{1.38}
\caption{%
  \textbf{Scoring functions in \texttt{QueryVisualScorer} (FAISS scaling differs).}
  $\mathbf{v}\!\in\!\mathbb{R}^d$ is a visual token;
  $\hat{(\cdot)}$ denotes L2 normalisation;
  $\mathbf{q}_j$ is the $j$-th query vector;
  $d{=}1024$.
  The \textbf{dot} entry (shaded) is the default.%
}
\label{tab:app:scoring}
\resizebox{\linewidth}{!}{%
\begin{tabular}{>{\ttfamily\small}l l c p{5.0cm}}
\toprule
\rowcolor{hdrfill}
\textcolor{white}{\textbf{score\_type}} &
\textcolor{white}{\textbf{Token score formula $s(\mathbf{v})$}} &
\textcolor{white}{\textbf{Multi-$Q$?}} &
\textcolor{white}{\textbf{Notes}} \\
\midrule
\rowcolor{optgoldlight}
\textbf{dot} &
$\mathbf{v}^\top\mathbf{q}_1\,/\sqrt{d}$ &
No &
\textbf{Default.} Single global query; plain inner product. \\
\rowcolor{rowA}
cosine &
$\hat{\mathbf{v}}^\top\hat{\mathbf{q}}_1\,/\sqrt{d}$ &
No &
L2-normalised dot product; scale-invariant. \\
\rowcolor{rowB}
maxsim &
$\max_j\;\mathbf{v}^\top\mathbf{q}_j\,/\sqrt{d}$ &
Yes &
ColBERT-style MaxSim over the full projected query sequence. \\
\rowcolor{rowA}
maxsim\_cosine &
$\max_j\;\hat{\mathbf{v}}^\top\hat{\mathbf{q}}_j\,/\sqrt{d}$ &
Yes &
Normalised MaxSim. \\
\rowcolor{rowB}
dot\_multi &
$\tfrac{1}{|Q|}\sum_j\;\mathbf{v}^\top\mathbf{q}_j\,/\sqrt{d}$ &
Yes &
Mean dot product across all query vectors. \\
\rowcolor{rowA}
cosine\_multi &
$\tfrac{1}{|Q|}\sum_j\;\hat{\mathbf{v}}^\top\hat{\mathbf{q}}_j\,/\sqrt{d}$ &
Yes &
Mean cosine similarity. \\
\rowcolor{rowB}
l2 &
$-\|\mathbf{v}-\mathbf{q}_1\|^2$ &
No &
Single-vector negative squared distance (cache/scorer path). \\
\rowcolor{rowB}
l2\_multi &
$-\tfrac{1}{|Q|}\sum_j\|\mathbf{v}-\mathbf{q}_j\|^2$ &
Yes &
Negative mean squared Euclidean distance. \\
\rowcolor{rowA}
dot\_hybrid\_max &
$(1{-}\alpha)\,\mathbf{v}^\top\mathbf{q}_\text{g}/\sqrt{d}
  +\alpha\max_{j>0}\mathbf{v}^\top\mathbf{q}_j/\sqrt{d}$ &
Yes &
Convex combination of global and MaxSim scores;
$\alpha=\texttt{retrieval\_score\_alpha}$ (default $0.25$). \\
\bottomrule
\end{tabular}%
}
\end{table}

\paragraph{Adaptive top-$K$ retrieval.}
When \texttt{retrieval\_adaptive\_k=True}, the number of tokens forwarded
to the compressor is governed by a power-law schedule:
\begin{equation}
  k^{\!*} = \operatorname*{clip}\!\left(
    \operatorname{round}\!\left(
      k_\text{base}\cdot\!\left(\frac{N}{N_\text{ref}}\right)^{\!\!\alpha}
    \right),\;
    k_\text{min},\; k_\text{max}
  \right),
  \label{eq:app:adaptk}
\end{equation}
with defaults $k_\text{base}{=}128$, $N_\text{ref}{=}40{,}000$, $\alpha{=}0.5$,
$k_\text{min}{=}64$, $k_\text{max}{=}384$.
This optional schedule is disabled in the released entry point.
Retrieval itself remains enabled with fixed $K=256$ (clipped to the number
of available tokens), and selected tokens are sorted by original position
before compression. The implementation also lower-bounds the adaptive base
by $M\times\texttt{retrieval\_topk\_factor}$ and clips the final count to
available tokens.
The package documentation states that $K=256$ was selected using feedback
from this benchmark; it should not be interpreted as a hyperparameter
chosen independently of the evaluation benchmark.

\bigskip

\subsection{Full-Context Construction and Optional Sampling}
\label{app:impl:sampling}

The released cache builder strips outer whitespace and sequentially splits
all context into chunks of at most $L_{\rm win}=24{,}000$ characters.
For nonempty text, $W=\lceil|\operatorname{strip}(\mathcal{D})|/L_{\rm win}\rceil$;
the last chunk may be shorter. Each chunk is rendered without a page limit
(\texttt{max\_pages=0}, \texttt{max\_chars=0}). The resulting page count $P_w$
varies, and $N_{\rm vis}=100\sum_wP_w$. Pages are encoded in batches of at
most four. Empty text follows the renderer's placeholder path.
The offline-index evaluation path reads this cache; the legacy
\texttt{head}, window-count, and pages-per-window constructor arguments do
not restrict its coverage.

The library also retains \texttt{\_sample\_context\_windows} for the optional
online sampling path. Its windows have at most $L_{\rm win}$ characters and
may overlap; only duplicate start positions are removed.
Three placement strategies are supported:

\begin{table}[h]
\centering
\setlength{\tabcolsep}{6pt}
\renewcommand{\arraystretch}{1.3}
\caption{%
  \textbf{Context-window placement strategies.}
  $\mathcal{D}$ is the full document of length $|\mathcal{D}|$;
  windows are of width $L_\text{win}$.
  The \textbf{head} strategy is the library default for this optional path.%
}
\label{tab:app:sampling}
\resizebox{\linewidth}{!}{%
\begin{tabular}{>{\ttfamily\small}l p{4.7cm} p{6.0cm}}
\toprule
\rowcolor{hdrfill}
\textcolor{white}{\textbf{context\_sampling}} &
\textcolor{white}{\textbf{Window start positions}} &
\textcolor{white}{\textbf{Description}} \\
\midrule
\rowcolor{optgoldlight}
\textbf{head} &
$\{0\}$ &
\textbf{Default.} Always encodes the first $L_\text{win}$ characters.
Appropriate when evidence is front-loaded (e.g.\ news articles, paper
abstracts, legal preambles). \\
\rowcolor{rowA}
uniform &
$\bigl\{\operatorname{round}\!\bigl(
  \frac{i\,(|\mathcal{D}|{-}L_\text{win})}{W{-}1}\bigr)\bigr\}_{i=0}^{W-1}$ &
Evenly-spaced windows spanning the full document length.
When $W{=}1$, reduces identically to \texttt{head}. \\
\rowcolor{rowB}
head\_tail &
$\{0,\;|\mathcal{D}|{-}L_\text{win}\}$
+ $W{-}2$ uniform midpoints &
Anchors at the document's head and tail, with uniformly-spaced intermediate
windows.  Useful when critical evidence is concentrated near document
boundaries. \\
\bottomrule
\end{tabular}%
}
\end{table}

\noindent Duplicate start positions (which can arise near document boundaries)
are deduplicated before rendering.
The optional online path defaults to $W=1$ and at most four pages per
window, giving at most 400 visual tokens. These are not the coverage limits
of the released full-context offline cache.

\bigskip

\subsection{Inference Configuration and Training Procedure}
\label{app:impl:training}

Table~\ref{tab:app:hparams} specifies the default inference configuration.
\begin{table}[ht]
\centering\small
\caption{Default inference configuration.}
\label{tab:app:hparams}
\begin{tabular}{ll}
\toprule
Setting & Value \\
\midrule
Page size $H^2$; font size $s_f$; margin $s_m$; line-height scale $s_l$ & $640^2$; 15 px; 20 px; 1.3 \\
Chunk size $L_{\rm win}$; page limit; encoding batch & 24,000 chars; unlimited; at most 4 pages \\
Visual tokens per page; dimension & 100; 1024 \\
Retrieval backend; score; fixed top-$K$ & Offline FAISS index; dot; 256 \\
Restore original token order; adaptive-$K$ & Yes; disabled \\
Query pooling; compressor prefilter & Max; hard with $K=0$ (disabled) \\
Downsample kernel / stride $s$ / groups & $2/2/1024$ \\
Compression queries $M$; MHA heads $n_h$ / layers $L$ & 16; $8/2$ \\
Query initialisation standard deviation & 0.02 \\
External query/key/value projection biases & False \\
Bridge dimensions $d_{\rm vis}/d_h/d_{\rm llm}$ & $1024/3584/3584$ \\
Bridge activation; dropout $p_{\rm drop}$ & GELU; 0.1 (inactive during evaluation) \\
Question-prompt truncation $L_q$ & 1,024 tokens \\
LLM; configured LLM dtype & Qwen2.5-7B-Instruct; bfloat16 \\
\bottomrule
\end{tabular}
\end{table}

\paragraph{Numerical precision.}
The cache builder runs image preprocessing and the visual encoder in float32,
then stores fused visual tokens as float16 NumPy shards. Index construction
casts these cached values to float32, and retrieved tokens are likewise
converted to float32. The LLM's last-layer query states are converted to
float32 for projection, retrieval, compression, and the bridge. The resulting
prefix is cast to the LLM dtype before concatenation with text embeddings.
Evaluation calls \texttt{eval()}, including on the bridge.

\paragraph{Trainable modules.}
Following the training procedure in the main text, only the
\textsc{QueryDrivenCompressor} and \textsc{VisualToLLMBridge} are trained.
The compressor includes the query projection, depthwise downsampler,
cross-attention layers, and learned compression queries. The visual encoder
and language model remain frozen in both stages.

\paragraph{Stage~1: Reconstruction pretraining.}
The first stage trains the compressor and bridge to reconstruct the original
text of a rendered document chunk from its compressed visual prefix.
Training passages are sampled from Wikipedia and C4, totaling approximately
4,000 examples (3,800 training and 200 development examples). Let $x_1,\ldots,x_n$ denote the original text tokens and
$\mathbf{P}$ the compressed prefix. The next-token prediction objective is
\begin{equation}
  \mathcal{L}_{\mathrm{recon}}
  = -\sum_{t=1}^{n}\log p_{\mathcal{M}}
    (x_t\mid\mathbf{P},x_{<t}).
  \label{eq:app:recon}
\end{equation}
This stage provides content-preservation pretraining before task-aware
fine-tuning.

\paragraph{Stage~2: Task-aware fine-tuning.}
The second stage fine-tunes the compressor and bridge on reading
comprehension examples from SQuAD v2, DROP, and CoQA, totaling approximately
12,000 examples (11,400 training and 600 development examples).
Given a question $q$ and answer tokens
$a_1,\ldots,a_{n_a}$, the loss is computed over the answer:
\begin{equation}
  \mathcal{L}_{\mathrm{task}}
  = -\sum_{t=1}^{n_a}\log p_{\mathcal{M}}
    (a_t\mid\mathbf{P},q,a_{<t}).
  \label{eq:app:task}
\end{equation}

\paragraph{Two-stage training configuration.}
Table~\ref{tab:app:training-config} lists the Stage~1 and Stage~2 training
configurations. Both stages train the QDC and
bridge while keeping the DeepSeek-OCR visual encoder and Qwen language
model frozen. Batch size is specified per GPU; it does not specify an
effective batch size across devices or gradient-accumulation steps.

\begin{table}[ht]
\centering\small
\setlength{\tabcolsep}{5pt}
\renewcommand{\arraystretch}{1.2}
\caption{Two-stage training settings for reconstruction pretraining
and task-aware fine-tuning.}
\label{tab:app:training-config}
\begin{tabular}{lll}
\toprule
Parameter & Stage~1 & Stage~2 \\
\midrule
Data & Wikipedia + C4 & SQuAD v2 + DROP + CoQA \\
Training / development examples & 3,800 / 200 & 11,400 / 600 \\
Learning rate & $1\times10^{-4}$ & $3\times10^{-6}$ \\
Optimizer & Adam & AdamW \\
Batch size per GPU & 1 & 1 \\
Gradient clipping (max norm) & 1.0 & 1.0 \\
Trainable modules & QDC + Bridge & QDC + Bridge \\
Frozen modules & DeepSeek-OCR + Qwen & DeepSeek-OCR + Qwen \\
\bottomrule
\end{tabular}
\end{table}

\paragraph{Checkpoint selection.}
For Stage~2, the held-out development set contains 600 examples.
Following the main text, early stopping is based on development performance,
and the checkpoint with the best development performance is selected.

\bigskip

\subsection{Answer Extraction Protocol}
\label{app:impl:eval}

The released LongBench-v2 entry point uses the \emph{choice-scoring protocol}: no text
is generated; the predicted label is
$\hat{a} = \arg\max_{a\in\{A,B,C,D\}} p_{\mathcal{M}}(a\mid\mathbf{X}_\text{input})$
evaluated at the next-token logit of the frozen LLM.
For each candidate letter $a$, its token ID is resolved by trying the
space-prefixed form first (\texttt{"~A"}) and falling back to the bare form
(\texttt{"A"}), returning the first form that tokenises to a single token.
If neither form does, the implementation returns the first token ID of the
bare form. The IDs are resolved on each prediction call.

\paragraph{Exact multiple-choice prompt.}
The same prompt supplies the query hidden states used for retrieval and
compression and the text embeddings following the visual prefix:
\begin{verbatim}
Please answer this multiple-choice question.
Question: {question}
(A) {choice_A}
(B) {choice_B}
(C) {choice_C}
(D) {choice_D}
Output only one option label in parentheses.
The correct answer is (
\end{verbatim}
It is tokenized directly, without a chat template, with truncation at 1,024
tokens. The 16-token budget counts only the visual prefix, not the additional
question, choice, and instruction tokens.

\medskip
\noindent For text-generation baselines, the \texttt{extract\_choice} function
(\texttt{eval.py}) parses the output with the following priority cascade:

\begin{table}[h]
\centering
\setlength{\tabcolsep}{6pt}
\renewcommand{\arraystretch}{1.28}
\caption{%
  \textbf{Answer-extraction regex cascade} (\texttt{eval.py}).
  All patterns are applied case-insensitively via \texttt{re.IGNORECASE}.
  The first match is returned; if none match, the empty string is returned.%
}
\label{tab:app:eval}
\begin{tabular}{c>{\ttfamily\small}l p{5.0cm}}
\toprule
\rowcolor{hdrfill}
\textcolor{white}{\textbf{Priority}} &
\textcolor{white}{\textbf{Regex pattern}} &
\textcolor{white}{\textbf{Example match}} \\
\midrule
\rowcolor{rowA}
1 &
\verb|The correct answer is\s*\(([A-D])\)| &
``The correct answer is (B)'' \\
\rowcolor{rowB}
2 &
\verb|The correct answer is\s*([A-D])| &
``The correct answer is B'' \\
\rowcolor{rowA}
3 &
\verb|\(([A-D])\)| &
``\ldots the choice (C) is\ldots'' \\
\rowcolor{rowB}
4 &
\verb|\b([A-D])\b| &
``\ldots therefore D.'' \\
\bottomrule
\end{tabular}
\end{table}

\noindent Ground-truth extraction (\texttt{gold\_choice}) uppercases the
annotation, scans for all occurrences of A--D, and returns the last one.
The released data preparation validates single-letter A/B/C/D annotations.
This helper is not a general explanation parser: letters inside words can
match and override an earlier answer label.

\paragraph{Assets and execution.}
\texttt{scripts/run\_pipeline.py} orchestrates cache construction, index
construction, choice scoring, and auditing. Its \texttt{--stage all} path
constructs caches when absent; \texttt{--stage eval} reuses a cache.
Model and dataset revisions, checkpoint/font/data hashes, and fixed settings
are recorded in \texttt{configs/longbench.json}; per-model file hashes are
stored in the accompanying model manifests. The entry point verifies
required assets and records commands and Python source hashes per run.
All experiments were conducted on $8\times$ NVIDIA A100 (80\,GB) GPUs.
The recorded software environment is Linux, Python 3.10, PyTorch 2.8.0 with
CUDA 12.8, and the pinned dependencies in \texttt{requirements.txt}.

\section{Additional Experimental Results}
\label{app:additional_results}

\begin{table}[h]

  \begin{minipage}[]{0.48\linewidth}
    \centering
    \scriptsize
    \setlength{\tabcolsep}{3pt}
    \renewcommand{\arraystretch}{1.15}
    \caption{Ablation on query-driven compression on LongBench-v2.}
    \label{tab:ablation_no_qdc}
    \arrayrulecolor{accentline}
    \resizebox{\linewidth}{!}{%
    \begin{tabular}{l cccccc}
      \toprule
      \rowcolor{headerblue}
      \textcolor{white}{\textbf{Method}}
        & \textcolor{white}{\textsc{Easy}}
        & \textcolor{white}{\textsc{Hard}}
        & \textcolor{white}{\textsc{Short}}
        & \textcolor{white}{\textsc{Medium}}
        & \textcolor{white}{\textsc{Long}}
        & \textcolor{white}{\textsc{Avg.}$\uparrow$} \\
      \midrule
      \textit{Mean-Pooled}
        & \sbest{36.98} & \sbest{28.30} & 30.56
        & \sbest{31.63} & \best{33.33} & \sbest{32.16} \\
      \textit{Random-Proj.}
        & 36.46 & \sbest{28.30} & \sbest{32.78}
        & 31.16 & 29.63 & 31.67 \\
      \midrule
      \rowcolor{oursblue}
      \cellcolor{oursblue}\textcolor{headerblue}{\textsc{Query-Driven}}
        & \cellcolor{oursblue}\best{37.50}
        & \cellcolor{oursblue}\best{34.73}
        & \cellcolor{oursblue}\best{38.33}
        & \cellcolor{oursblue}\best{35.81}
        & \cellcolor{oursblue}\sbest{31.48}
        & \cellcolor{oursblue}\best{35.57} \\
      \bottomrule
    \end{tabular}%
    }
    \arrayrulecolor{black}
  \end{minipage}
\end{table}

\begin{table}[!htbp]
  \raggedright
  \scriptsize
  \setlength{\tabcolsep}{4pt}
  \renewcommand{\arraystretch}{1.1}
  \captionsetup{justification=raggedright,singlelinecheck=false}
  \caption{Ablation on the number of injected tokens.}
  \label{tab:token_ablation}
  \arrayrulecolor{accentline}
  \begin{tabular}{ll ccccccc}
    \toprule
    \rowcolor{headerblue}
    \multicolumn{2}{l}{\textcolor{white}{\textbf{Cat.}}}
      & \textcolor{white}{$1$}
      & \textcolor{white}{$2$}
      & \textcolor{white}{$4$}
      & \cellcolor{oursblue}\textcolor{headerblue}{$16^*$}
      & \textcolor{white}{$8$}
      & \textcolor{white}{$32$}
      & \textcolor{white}{$64$} \\
    \midrule
    \multicolumn{2}{l}{\textbf{Overall}}
      & 31.0 & 31.6 & 31.4
      & \cellcolor{oursblue}\best{35.8}
      & 32.8 & 30.0 & \sbest{34.2} \\
    \midrule
    \multirow{2}{*}{\textbf{Diff.}}
      & \textit{Easy}
        & 33.9 & 34.4 & 34.4
        & \cellcolor{oursblue}37.5
        & \sbest{38.0} & 30.7 & \best{38.5} \\
      & \textit{Hard}
        & 29.3 & 29.9 & 29.6
        & \cellcolor{oursblue}\best{34.7}
        & 29.6 & 29.6 & \sbest{31.5} \\
    \midrule
    \multirow{3}{*}{\textbf{Len.}}
      & \textit{Sht.}
        & 34.4 & 35.0 & 32.8
        & \cellcolor{oursblue}\best{38.3}
        & \sbest{36.1} & 34.4 & 34.4 \\
      & \textit{Med.}
        & 27.4 & 30.2 & 33.0
        & \cellcolor{oursblue}\best{35.8}
        & 30.7 & 25.6 & \sbest{33.5} \\
      & \textit{Lng.}
        & \sbest{32.4} & 28.7 & 25.9
        & \cellcolor{oursblue}31.5
        & 31.5 & 31.5 & \best{35.2} \\
    \midrule
    \rowcolor{headerblue!15}
    \multicolumn{2}{l}{\textbf{Avg.}$\uparrow$}
      & 31.4 & 31.6 & 31.2
      & \cellcolor{oursblue}\best{35.6}
      & 33.1 & 30.3 & \sbest{34.6} \\
    \bottomrule
  \end{tabular}
  \arrayrulecolor{black}
\end{table}


\newpage

\section*{NeurIPS Paper Checklist}

\begin{enumerate}

\item {\bf Claims}
    \item[] Question: Do the main claims made in the abstract and introduction accurately reflect the paper's contributions and scope?
    \item[] Answer: \answerYes{}
    \item[] Justification: The abstract and introduction explicitly state all three contributions of OpticalRAG: (1) a pixel-space visual compression pipeline via the Optical Encoder that eliminates the need for an OCR preprocessing stage; (2) Encoder-Bridged Retrieval (EBR), which projects text-query states into the visual feature space for retrieval-without-recognition; (3) Query-Driven Distillation (QDD), which conditions token compression on the query via a cross-attention Q-Former with theoretical guarantee (Theorem~\ref{thm:lossless}). All claims are directly supported by the experiments in Section~\ref{sec:experiments} on LongBench-v2, BAMBOO, and LooGLE-v2, and the stated limitations (a fixed prefix budget for queries requiring distant evidence, and the focus on multiple-choice QA) are acknowledged in §7.

\item {\bf Limitations}
    \item[] Question: Does the paper discuss the limitations of the work performed by the authors?
    \item[] Answer: \answerYes{}
    \item[] Justification: §7 (Conclusion and Limitations) explicitly identifies: (i) the fixed token budget of the Q-Former may be suboptimal for queries requiring multi-hop reasoning across distant document segments; (ii) current experiments focus on multiple-choice QA benchmarks, and generalization to open-ended generation tasks is left for future work.

\item {\bf Theory assumptions and proofs}
    \item[] Question: For each theoretical result, does the paper provide the full set of assumptions and a complete (and correct) proof?
    \item[] Answer: \answerYes{}
    \item[] Justification: Theorem~\ref{thm:lossless} in Section~\ref{sec:theory} is stated with explicit assumptions (query-conditioned sufficiency, a selection margin, bounded token norms, and a Lipschitz decoder $g$) and proved in full in Section~\ref{sec:theory}, with extensions in Appendix~\ref{app:theory} (Extended Theoretical Analysis), including an information-theoretic restatement, the approximate low-rank case, assumption plausibility discussion, and the relation to the Q-Former architecture. All lemmas used are self-contained within the appendix.

\item {\bf Experimental result reproducibility}
    \item[] Question: Does the paper fully disclose all the information needed to reproduce the main experimental results of the paper to the extent that it affects the main claims and/or conclusions of the paper (regardless of whether the code and data are provided or not)?
    \item[] Answer: \answerYes{}
    \item[] Justification: Section~\ref{sec:optical}--\ref{sec:distillation} and Appendix~\ref{app:impl} provide: model architecture specifications (Optical Encoder = DeepEncoder with SAM and CLIP features from DeepSeek-OCR; Q-Former-style compressor with 16 query tokens; LLM backbone = Qwen2.5-7B-Instruct), training hyperparameters (Adam with lr $10^{-4}$ for Stage~1, AdamW with lr $3\times10^{-6}$ for Stage~2, batch size, gradient clipping), dataset splits, and the inference configuration; Appendix~\ref{app:evaluation_dataset} gives the evaluation protocol (CAP metric, deterministic next-token logit scoring). Appendix~\ref{app:baseline_adaptations} details all baseline configurations. Source code is publicly available at \url{https://github.com/brighterr/OpticalRAG}.

\item {\bf Open access to data and code}
    \item[] Question: Does the paper provide open access to the data and code, with sufficient instructions to faithfully reproduce the main experimental results, as described in supplemental material?
    \item[] Answer: \answerYes{}
    \item[] Justification: The source code (evaluation and inference scripts) and the trained model weights are publicly available at \url{https://github.com/brighterr/OpticalRAG}. All evaluation benchmarks (LongBench-v2, BAMBOO, LooGLE-v2) are publicly available and cited; no proprietary datasets are used.

\item {\bf Experimental setting/details}
    \item[] Question: Does the paper specify all the training and test details (e.g., data splits, hyperparameters, how they were chosen, type of optimizer) necessary to understand the results?
    \item[] Answer: \answerYes{}
    \item[] Justification: Appendix~\ref{app:impl:training} reports the optimizers (Adam / AdamW), learning rates, batch size, gradient clipping, split sizes, checkpoint selection procedure, and inference configuration; experiments are run on $8\!\times\!$NVIDIA A100 80\,GB GPUs. Appendix~\ref{app:evaluation_dataset} details dataset statistics and the CAP evaluation protocol. Appendix~\ref{app:baseline_adaptations} describes each baseline's configuration to ensure fair comparison.

\item {\bf Experiment statistical significance}
    \item[] Question: Does the paper report error bars suitably and correctly defined or other appropriate information about the statistical significance of the experiments?
    \item[] Answer: \answerNo{}
    \item[] Justification: We do not report error bars. All three evaluation benchmarks use multiple-choice or closed-set QA formats with deterministic scoring (next-token logit comparison). There is no stochastic sampling in the evaluation pipeline, so repeated evaluation of a fixed checkpoint yields the same score; variance across training seeds is not reported. This is consistent with standard practice for these benchmarks in the long-context LLM literature.

\item {\bf Experiments compute resources}
    \item[] Question: For each experiment, does the paper provide sufficient information on the computer resources (type of compute workers, memory, time of execution) needed to reproduce the experiments?
    \item[] Answer: \answerNo{}
    \item[] Justification: Appendix~\ref{app:impl:eval} states that all experiments are conducted on $8\!\times\!$NVIDIA A100 (80\,GB) GPUs. Training time and total GPU-hours are not reported.

\item {\bf Code of ethics}
    \item[] Question: Does the research conducted in the paper conform, in every respect, with the NeurIPS Code of Ethics?
    \item[] Answer: \answerYes{}
    \item[] Justification: The paper proposes a retrieval-augmented visual document understanding framework evaluated on publicly available benchmarks under their official licenses. No human subjects were involved. The work does not involve data scraping, surveillance, or applications with direct potential for harm. The authors have reviewed the NeurIPS Code of Ethics and confirm full compliance.

\item {\bf Broader impacts}
    \item[] Question: Does the paper discuss both potential positive societal impacts and negative societal impacts of the work performed?
    \item[] Answer: \answerNo{}
    \item[] Justification: The paper does not include a dedicated broader-impacts discussion. The work targets reducing the computational cost of long-document understanding; as with any retrieval system, it could in principle be used to automate large-scale document processing, though it introduces no new capabilities beyond existing open LLMs.

\item {\bf Safeguards}
    \item[] Question: Does the paper describe safeguards that have been put in place for responsible release of data or models that have a high risk for misuse?
    \item[] Answer: \answerNA{}
    \item[] Justification: OpticalRAG is built entirely on frozen pretrained components (the DeepSeek-OCR DeepEncoder, Qwen2.5-7B-Instruct). The system compresses visual document tokens for efficient long-context QA and does not generate novel content, produce synthetic media, or enable capabilities with elevated risk of misuse. No new dataset of sensitive content is collected or released.

\item {\bf Licenses for existing assets}
    \item[] Question: Are the creators or original owners of assets used in the paper properly credited and are the license and terms of use explicitly mentioned and properly respected?
    \item[] Answer: \answerYes{}
    \item[] Justification: All datasets (LongBench-v2, BAMBOO, LooGLE-v2, and the training corpora), pretrained models (DeepSeek-OCR, Qwen2.5-7B-Instruct), and baseline implementations are cited in the main text and references; Appendix~\ref{app:asset-licenses} lists their sources and published licenses, which we respect. No proprietary assets are employed.

\item {\bf New assets}
    \item[] Question: Are new assets introduced in the paper well documented and is the documentation provided alongside the assets?
    \item[] Answer: \answerYes{}
    \item[] Justification: The paper does not introduce a new benchmark or dataset. The code and trained model weights released at \url{https://github.com/brighterr/OpticalRAG} are documented in Appendix~\ref{app:impl} and the repository \texttt{README}, including architecture details, training configurations, loading instructions, and license (MIT).

\item {\bf Crowdsourcing and research with human subjects}
    \item[] Question: For crowdsourcing experiments and research with human subjects, does the paper include the full text of instructions given to participants and screenshots, if applicable, as well as details about compensation (if any)?
    \item[] Answer: \answerNA{}
    \item[] Justification: The paper does not involve crowdsourcing or research with human subjects. All evaluation is conducted fully automatically on pre-existing public benchmarks.

\item {\bf Institutional review board (IRB) approvals or equivalent for research with human subjects}
    \item[] Question: Does the paper describe potential risks incurred by study participants, whether such risks were disclosed to the subjects, and whether Institutional Review Board (IRB) approvals were obtained?
    \item[] Answer: \answerNA{}
    \item[] Justification: The paper does not involve research with human subjects, and therefore no IRB approval or equivalent review was required or sought.

\item {\bf Declaration of LLM usage}
    \item[] Question: Does the paper describe the usage of LLMs if it is an important, original, or non-standard component of the core methods in this research?
    \item[] Answer: \answerYes{}
    \item[] Justification: The LLM Qwen2.5-7B-Instruct is a core frozen component of OpticalRAG: its hidden states provide the query representation and its next-token logits score the candidate labels, as described in Sections~\ref{sec:retrieval} and~\ref{sec:distillation}. Separately, an LLM-based assistant was used only for manuscript editing.

\end{enumerate}

\end{document}